\documentclass[10pt,twocolumn,letterpaper]{article}

\usepackage{iccv}
\usepackage{times}
\usepackage{epsfig}
\usepackage{graphicx}
\usepackage{amsmath}
\usepackage{amssymb}

\usepackage{caption}
\usepackage{xspace}
\usepackage[subrefformat=parens]{subcaption}

\usepackage[utf8]{inputenc} 
\usepackage{url}            
\usepackage{booktabs}       
\usepackage{amsfonts}       
\usepackage{nicefrac}       
\usepackage{microtype}      
\usepackage{xcolor}         

\usepackage{amsmath}
\usepackage{amssymb}
\usepackage{mathtools}
\usepackage{amsthm}
\usepackage{algorithm}
\usepackage{algorithmic}

\usepackage{mathtools}
\usepackage{amsthm}

\theoremstyle{plain}
\newtheorem{theorem}{Theorem}[section]

\theoremstyle{definition}
\newtheorem{definition}[theorem]{Definition}

\theoremstyle{remark}

\newcommand{\SieveStreaming}{{\textsc{Sieve-Streaming}}\xspace}
\newcommand{\SSPP}{{\textsc{Sieve-Streaming}}{\texttt{++}}\xspace}
\newcommand{\ThreeSieves}{{ThreeSieves}\xspace}

\newcommand{\TheTitle}{Active Learning for Deep Neural Networks on Edge Devices}

\usepackage[pagebackref=true,breaklinks=true,letterpaper=true,colorlinks,bookmarks=false]{hyperref}

\iccvfinalcopy 

\def\iccvPaperID{****} 

\ificcvfinal\fi

\begin{document}

\title{\TheTitle}

\author{\qquad Yuya Senzaki \qquad Christian Hamelain\\
Idein inc.\\
{\tt\small \{senzaki, christian\}@idein.jp}\\
}

\maketitle
\ificcvfinal\thispagestyle{empty}\fi

\begin{abstract}
  When dealing with deep neural network (DNN) applications on edge devices,
  continuously updating the model is important.
  Although updating a model with real incoming data is ideal,
  using all of them is not always feasible due to limits, such as
  labeling and communication costs.
  Thus, it is necessary to filter and select the data to use for training (i.e., active learning) on the device.
  In this paper, we formalize a practical active learning problem for DNNs on edge devices
  and propose a general task-agnostic framework to tackle this problem, which reduces it to a stream submodular maximization.
  This framework is light enough to be run with low computational resources,
  yet provides solutions whose quality is theoretically guaranteed thanks to the submodular property.
  Through this framework, we can configure data selection criteria flexibly,
  including using methods proposed in previous active learning studies.
  We evaluate our approach on both classification and object detection tasks
  in a practical setting to simulate a real-life scenario.
  The results of our study show that the proposed framework outperforms all other methods in both tasks,
  while running at a practical speed on real devices.
\end{abstract}

\section{Introduction}
\label{sec:introduction}

With advances in both hardware and software domains,
running inference of deep neural networks (DNNs) on edge devices is now practical,
even for relatively cheap devices~\cite{sandler2018mobilenetv2, howard2019searching, tan2019mnasnet}.
Although these devices have lower computation resources and storage than servers,
inference on these devices has many advantages over inference on servers;
as it does not require sending input data (e.g., image)
over the network,
the latency becomes low, communication cost is decreased, and privacy is protected.

As DNNs are considerably large and require large computation for training~\cite{tan2019efficientnet},
a model for the edge is usually trained on a server and subsequently deployed to the device.
In this setting, it is unlikely that the model constructed in the beginning is ``perfect'' for two reasons.
One is a problem of data.
While deep learning is often very greedy for data~\cite{sun2017revisiting},
the gathering and labeling of data are time-consuming and expensive tasks.
Moreover, the required amount and quality of data
for acquiring the desired accuracy is unknown.
Thus preparing data in sufficient amounts before training is challenging.
Another reason is the problem known as \textit{domain shift}.
When the model is deployed after training on a server,
depending on its deployment site, the distribution of input data may differ from
that of training data due to several factors, such as temporal or environmental ones;
Ultimately, constructing a model robust for these variations is ideal.
For edge devices, however, this is unrealistic since such an adaptive model would require considerable memory or computational resources.

\begin{figure*}[!t]
  \centering
  \includegraphics[keepaspectratio, width=.88\textwidth]{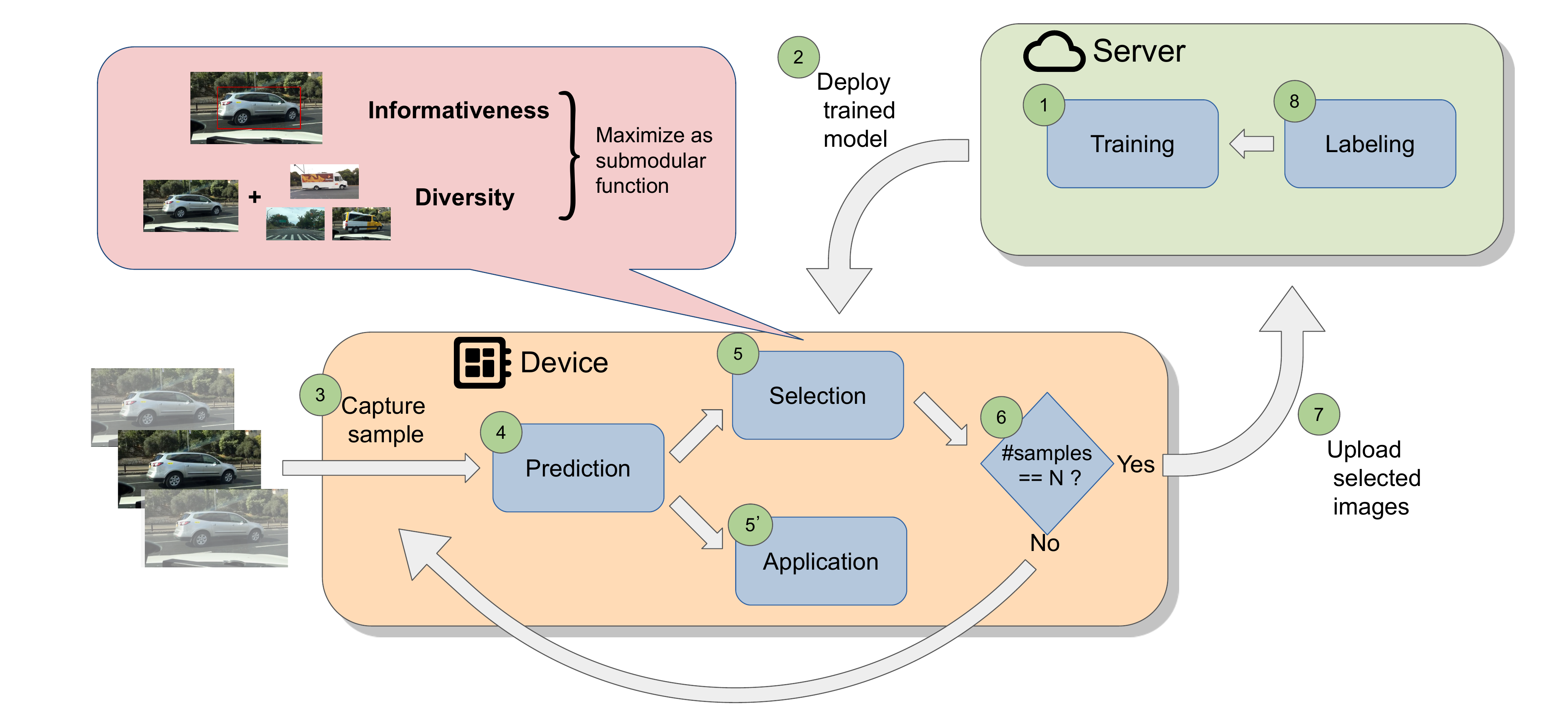}
  \caption{
    Overview of the whole workflow.
    1) Train a model with labeled data.
    2) Deploy / update trained model to the device.
    3) Capture one sample as input.
    4) Run prediction.
    5) Decide whether store or discard sample by considering both informativeness and diversity.
    5') Use the result of prediction.
    6) Repeat 3--6 $N$ times.
    7) Send selected data to server.
    8) Annotate received data.
  }
  \label{fig:overview}
\end{figure*}

For these reasons, continuous updating of the model is preferable to one-shot training.
In this scenario, the model also benefits from having real input data to train with (i.e., the training data distribution is the same as the target test data).
Meanwhile, it is generally unrealistic to send all incoming data to a server.
The amount of data will rapidly become too huge, especially if the processing is performed at a high frequency.
At this time, the cost induced by the communication to the remote server is prohibitive,
especially if the connection fee is on a pay-as-you-go basis.
Furthermore, almost all data will contain a considerable number of similar samples,
considering chronologically consequent data.
For these reasons, we need to perform data selection on the device.

Selecting data and training with it is known as \textit{active learning},
which is a well studied field~\cite{settles2009active}.
Thus, in this paper we consider the problem we face as \textit{active learning on edge devices}.
Unfortunately, however, many approaches proposed in the literature are unable to
trivially extend to the scenario we consider:
as discuss in Section~\ref{sec:problem_definition}, it is natural to assume that there are more constraints on edge devices.

In this paper, we first discuss the properties that reflect the real-world constraints and formalize the novel problem,
\textit{active learning on edge devices}.
The constraints of this problem are manifold, making it difficult to apply existing active learning methods.
Thus we propose a task-agnostic framework for solving this problem.
In this framework, we reduce the problem to a submodular maximization satisfying those constraints.
As a result, we can solve the problem with theoretical guarantees on solution optimality by this framework.
We show an overview in Figure~\ref{fig:overview}.
In the empirical evaluations, we show actual constructions for different tasks
and demonstrate that our approach outperforms all other methods.
In addition, we also show that our framework can be used practically through evaluations on a real device.

\section{Related work}
\label{sec:related_work}

\paragraph{Active Learning}

Active learning aims to train a model with few labeled samples
by selecting data to label from a (large) pool of unlabeled data~\cite{settles2009active}.
This field has been studied for a long time, and since larger
amounts of data became available, its importance further increased.
In a standard active learning setup, data is selected from a set of candidates
and then fed to an \textit{oracle} (e.g., a human annotator) that
returns annotation.
The model is trained with the labeled data, and this process is typically iterated
multiple times (we call each iteration a \textit{round}).
It is assumed that labeling is costly; this is
the primary motivation for reducing the amount of data to label.

While it is common to choose data from candidates,
two settings exist depending on the accessibility of candidate data:
\textit{pool-based} and \textit{stream-based} setting~\cite{settles2009active}.
The former allows access to all candidate data, and it is the most popular setting.
In the latter setting, we can only look at each sample once in a determined order and keep
only a (small) subset of the data.
Therefore, in the pool-based setting, we can consider the relationship between
arbitrary data in the ground set, while it is limited in stream-setting.
Moreover, we distinguish two policies by the amount of data
sent to an oracle at a time; \textit{one-by-one} and \textit{batch-mode}.
For the one-by-one strategy, we only need to consider how to evaluate each sample solely,
such as uncertainty~\cite{joshi2009multiclass, gal2017deep, wang2021neural},
expected model output change~\cite{freytag2014selecting, settles2008multipleinstance}
or expected error reduction~\cite{roy2001optimal}.
While Amin et al.~\cite{amin2020understanding} showed that these methods can be simply extended to batch-mode,
they still evaluate each sample independently
(see Section~\ref{sec:problem_definition} to further discussion about batch-mode).
However, it is known that
these kinds of approaches are ineffective for DNNs~\cite{sener2018active}.
A batch-mode method, which has to take both informativeness and diversity into account, is thus required.

Various approaches are proposed to select a diversified batch of data.
Sener and Savarese~\cite{sener2018active} formalizes the active learning problem
as a core-set selection in the image space.
Ash et al.~\cite{ash2020deep} proposes to create diversity through probabilistic sampling
with probability based on the distance between samples.
Kirsche et al.~\cite{kirsch2019batchbald} considers the mutual information between a batch and model parameters,
although their target is Bayesian Neural Networks.
Since submodular functions, which are described later, naturally model the diversity,
some works reduce active learning
to submodular optimization~\cite{hoi2006batch, wei2015submodularity, kaushal2019learning}.
However, these methods are pool-based and cannot easily be applied to the stream setting.
As described above, the stream setting compels us to consider relationship information,
making the batch-mode active learning in this setting challenging.

\paragraph{Submodular Maximization}
If a set function $f: 2^V \rightarrow \mathbb{R}$ has the following
\textit{diminishing returns} property, it is called \textit{submodular function}:
$f(S \cup \{e\}) - f(S) \geq f(T \cup \{e\}) - f(T)$
where $S \subseteq T \subseteq V$ and $e \in V \backslash T$.
Since this property is similar to concavity,
various problems can reduce to finding the set maximizing $f$,
i.e., a submodular maximization problem~\cite{krause2012submodular}.
There are many variations of this problem, depending on the constraints or the nature of $f$.
One of the most popular settings is when $f$ is a non-negative monotone submodular function, and
the size of the set is bounded.
Here, submodular function $f$ is \textit{monotone}
if for all $S \subseteq T$ we have $f(S) \leq f(T)$.
Although this problem is NP-hard, it is proven that simple greedy algorithm guarantees
$(1-1/e)$ approximation ratio, i.e., the worst-case solution is greater than
$(1-1/e)$ times the optimal solution~\cite{nemhauser1978analysis}.  
This is a promising approach when we can access all samples; however,
recent applications, including data mining and machine learning, require treating large amounts
of data and/or sequentially obtained data.
To solve the submodular maximization problem in this stream setting,
several algorithms have been proposed.
Badanidiyuru et al.~\cite{badanidiyuru2014streaming} proposes \SieveStreaming,
which is the first one-pass streaming algorithm.
This algorithm has a $(1/2 - \epsilon)$ approximation ratio with $O(K\log(K)/\epsilon)$ memory.
After this work, the improved algorithms \SSPP~\cite{kazemi2019submodular} and
\ThreeSieves~\cite{buschjager2020very} were proposed.
The former improves memory usage to $O(K/\epsilon)$ while keeping the same approximation ratio,
and the latter decreases it to $O(K)$ but makes the lower bound guarantee probabilistic.
We provide an overview of the algorithms in Appendix~\ref{appendix:algo_overview}.
In the experiments (Section~\ref{sec:experiments}), we use \SSPP to solve the problem our framework formalizes.

\section{Active learning for DNNs on edge devices}
\label{sec:problem_definition}

\subsection{Setting}

In this paper, we consider a situation where a DNN model is running on a device
and continuously performing inference on input data (e.g., from a camera, a microphone, among others).
The model is trained on a server and deployed to the device.
Our aim is to select data useful for training from the incoming data flow, without stopping inference.
When simultaneously running the prediction and selection,
we assume the inference and the screening of one sample must be completed
before processing the next sample.

To consider common devices such as Raspberry Pi and NVIDIA Jetson,
we assume that the available storage space is largely smaller than the total incoming data size.
Since one of the advantages of these devices is their relatively low cost,
we do not consider the use of extended storage space.
In addition, we also assume that the amount of data that can be sent is limited.
As described in Section~\ref{sec:introduction}, it is natural to assume that the communication cost cannot be ignored.
In this setting, we cannot take the \textit{pool-based} approach.
To solve the problem via this approach, it would require to store all the data or send it to the server.
Since this is not possible due to the above-mentioned reasons,
we have to consider the \textit{stream setting}, where data arrive continuously, and
only a part of them can be stored.

As a deep learning context, we also have to consider batch-setting, as prior works mentioned~\cite{sener2018active, ash2020deep}.
In this setting, multiple data samples are selected in one round,
in contrast to the traditional method of selecting one per round.
With this, one may have to consider correlations among multiple samples rather than treating each sample solely.

Consequently, we can regard the problem as stream-based batch-mode active learning.
There are mainly two types of batch-mode method regarding the terminating condition:
1) greedily select and send to oracle as soon as the batch reaches a given size \cite{amin2020understanding}
or 2) send the batch of data to the oracle after seeing a given number of samples.
While the former has the potential to reduce unnecessary selection,
the amount of data selected per hour is unpredictable at a given sample frequency
and it would not be preferable in many practical applications.
Thus, we consider the latter setting.
Under these assumptions,
we need to consider the nature of data, which is different from the usual active learning setting.
Moreover, we consider two scenarios that reflect problems faced in practical applications.
These are closely related to the evaluation of the method in this problem.

\paragraph{Peculiarity of data}
To the best of our knowledge, most active learning studies consider data that is in some sense filtered, i.e.,
it is (implicitly) assumed that every candidate is a useful data sample for training.
For example, when selecting data from an existing dataset,
almost all samples are presumably relevant (except for mislabeling, etc.).
However, when we consider data selection on a device,
it is natural to assume that the data will be in various conditions
such as unclear, blurred, redundant, among others.
We generally classify those into the three following statuses: ``imbalanced'', ``duplicated'', and ``unuseful''.
The first one indicates class imbalance, i.e.,
some classes of data rarely appear, while instances of other classes often appear.
The second condition is when almost identical samples exist in candidates.
If the model runs at a high speed, the frequency of data capture is also high.
Then, a data sample would be often similar to its surrounding samples.
The last status describes the presence of data such as ``non-object'' or ``background'' samples.
For example, in a face detection task, images without a face in it are not useful for increasing
accuracy.
We have to consider those peculiarities in the evaluation.
We show examples from real data in Appendix (Figure~\ref{fig:data_peculiarity}).

\paragraph{Scenarios}
With regard to active learning on edge devices,
we consider two main scenarios: \textit{cold-start} and \textit{domain adaptation}.
Although both scenarios predict and gather data in parallel,
the former scenario considers training the model with a small initial dataset.
This makes deployment earlier; consequently, the total operating time increases.
The latter is a scenario where
the model is initially trained with a dataset, which is usually not small, but
its distribution is different from the target environment
and is later updated with the collected data.
This is intended to treat the domain shift, as mentioned above.
There is a similar problem setting called \textit{unsupervised domain adaptation}~\cite{ganin2016domainadversarial, long2015learning},
but this is different in that no label of target domain data is available.
In both scenarios, the goal is to adapt the model to its deployment environment.

\subsection{Problem Definition}

Let $\mathcal{X}$ be the instance space,
$V \subseteq \mathcal{X}$ whole incoming data set (i.e., ground set).
Let ${V_n, n \in \{1,\dots,N\}}, V_n \subseteq V$  be a partition of $V$ by round $n$.
We aim to select at each round $n$ a subset $S_n \subseteq V_n$,
$|S_n| \leq K$, of data to train a model, where $K$ is a given constant.
Since we consider the stream setting, we cannot access all $x \in V_n$ at the same time.
Explicitly, at step $t \in \{1,\dots,|V_n|\}$ in round $n$,
we have to decide whether store or discard a sample $x_t \in V_n$ as soon as it arrives.
At this time, we can evaluate only stored samples and $x_t$.
Moreover, we make no assumption on the independence between samples, i.e., samples are not \textit{independent and identically distributed (iid)}.
In a typical stream (e.g., a video stream), there is indeed a temporal correlation between following samples,
and therefore samples are not independent.

As an active learning problem, we assume we have an initial dataset
$D_0$ at the beginning.
For each round, after selecting $S_n$, we label all samples in $S_n$,
add them to the dataset (i.e., $D_n = D_{n-1} \cup S_n$), and retrain the model with $D_n$.
Let a model parameter be $\theta$ and its loss function be $l$.
Depending on the distribution where the samples in $V$ and $D_0$ are drawn from,
we define the following two scenarios.
Note that in both scenarios, there is no constraint on the nature of the loss function $l$, i.e.,
we can consider any supervised tasks.
\begin{definition}[Cold-start Scenario]
  Given $D_0$, whose samples are drawn from the same distribution as $V$.
  Let this distribution be $\mathcal{P}$ and $|D_0| \ll |V|$.
  In this scenario, the goal is minimizing $\mathbb{E}_{z \sim \mathcal{P}} \ l(z ; \theta)$.
\end{definition}

\begin{definition}[Domain Adaptation Scenario]
  Given $D_0$, whose samples are drawn from a different distribution than $V$.
  Let these distributions be $\mathcal{S}$ and $\mathcal{T}$ respectively.
  In this scenario, the goal is minimizing $\ \mathbb{E}_{z \sim \mathcal{T}} \ l(z ; \theta)$.
\end{definition}

While we assume the existence of an initial dataset in this paper,
its construction would be a good research topic.
Especially in the cold-start scenario,
one may consider collecting data randomly or only with a model-agnostic method using the device, and create from scratch the $D_0$ dataset.
This collection method could be inspired by the approach we propose.

\section{Approach}
\label{sec:approach}

\subsection{Active Learning Framework}
Let $f$ be a set function $f: 2^V \rightarrow \mathbb{R}$.
We define $f(S)$ representing a valueness of $S$.
The goal is to select subset $S \subseteq V, |S| \leq K$ which maximize $f$.
To meet the conditions of a stream setting, the computation of $f(S)$ must depend only on the elements of the set $S$.
Here, we divide the valueness of $S$ into the informativeness and diversity:
the informativeness represents the total contribution of usefulness for each sample of the batch taken solely.
Generally, the sum of each samples' informativeness is the upper bound of the informativeness of a set
and it is smaller when there are some dependencies between samples.
To compensate for this, we introduce the diversity, which measures the value of the batch
as a whole considering relation between samples, e.g., representing how a batch contains little overlapping.
Any valueness of a sample due to the presence of another sample in the batch, i.e. relationship between samples, is considered
in the diversity criterion.
These criteria are expected to deal with the peculiarities of data described in Section~\ref{sec:problem_definition}:
informativeness will be high for rare-class and useful data, and diversity will be high if samples are less duplicated.

To consider these criteria simultaneously,
we define the objective function $f$ as the linear combination of these:
\begin{equation}
  \label{eq:objective_function}
  f(S) = {\lambda}_i \cdot f_i(S) + {\lambda}_d \cdot f_d(S)
\end{equation}
where $f_i, f_d: 2^V \rightarrow \mathbb{R}$ are respectively the function for informativeness and diversity,
and ${\lambda}_i, {\lambda}_d \in \mathbb{R}^{+}$ are hyperparameters for balancing.
To obtain solutions with worst-case guarantees,
we aim to maximize $f(S)$ in a submodular optimization manner.
Then, this formalization is convenient:
if both $f_i$ and $f_d$ are monotone submodular, $f$ is also monotone submodular.
Thus the policy is to construct each function as a monotone submodular function.
In the following, we describe how we design each function.

\paragraph{Informativeness}
Quantifying how informative a sample is is a task-dependent problem.
For example, softmax entropy~\cite{settles2009active} may be a good candidate for classification tasks,
but is insufficient for object detection tasks~\cite{girshick2014rich, liu2016ssd}
since it does not reflect the results of predicted bounding boxes.
However, regardless of the task, the computation of each sample's informativeness can be done independently.
Thus we define $f_i$ by using a function $g: \mathcal{X} \rightarrow \mathbb{R}^{+}$
and denote as
\begin{equation}
  \label{eq:informativeness_function}
  f_i(S) = \sum_{x \in S} g(x).
\end{equation}
This is a form of linear modular function, and as $\forall x, g(x) \geq 0$ it is a monotone submodular function, thus fulfills the condition.
The benefit of this formalization is that we can
utilize various functions proposed in the previous active learning studies~\cite{settles2009active, kao2019localizationaware} as $g$,
since many focus on the same problem, i.e., how to quantify the sample informativeness.
This means there is room to choose the best function for the target task.
We show concrete examples of constructed function in Section~\ref{sec:experiments}.

\paragraph{Diversity}
In contrast to the informativeness, we have to consider the relationships of samples in a batch,
i.e., it cannot be represented as the sum of scores by sample.
Usually, this relation is defined through similarity, i.e.,
high diversity means low similarity between samples.
While there are many ways to represent the similarity in a set,
designing it under the submodular constraint is not trivial.
Therefore, constructing the diversity function of a set is
summed up in two points: how we evaluate (dis)similarity between two samples and
how we translate it into diversity.

As described in Section~\ref{sec:related_work}, expressing diversity is studied in various contexts.
However, in most works, the proposed measure does not satisfy the constraints we consider, e.g.,
BADGE~\cite{ash2020deep} requires a ground set for representing the diversity of one subset.
Utilizing log-determinant~\cite{lawrence2003fast, kulesza2012determinantal} allows for flexible
 design while satisfying constraints.
Considering the similarity matrix
$M = [k(x_i, x_j)]_{ij}$
for all $x \in V$
where $k: \mathcal{X} \times \mathcal{X} \rightarrow \mathbb{R}$ is a similarity function.
Note that $M$ is always a $|V| \times |V|$ symmetric matrix, and all the diagonal values
are equal to its largest value by construction.
Considering the problem with combinatorial optimization perspective,
let $M_S$ be the sub-matrix of $M$ indexed by $S$, i.e., $M_S = [k(x_i, x_j)]_{ij}$ for $x \in S$.
Then, the diversity is described as
\begin{equation}
  \label{eq:diversity_function}
  f_d(S) = \frac{1}{2} \log \det (I + \alpha M_S)
\end{equation}
where $I$ is the identity matrix and $\alpha \in \mathbb{R}^{+}$ is a scaling parameter.
Here, what we are interested in is what kind of functions we can use for $k$.
Since similarity metrics have task- and context-dependent aspects 
(e.g., pictures of a bird and an airplane may be similar in image space but are not
in terms of meaning),
we want to allow various functions.
Fortunately, we can use any positive semidefinite kernel~\cite{hofmann2008kernel} as a similarity function.
It is known that if $X \in \mathbb{R}^{n \times n}$ is symmetric and its smallest eigenvalue is
larger than 1, the function $f(S) = \log \det (X_S)$ is monotone submodular~\cite{sharma2015greedy}.
Thus if $k$ is positive semidefinite (i.e. $M$ is also positive semidifinite),
$f_d$ in Eq.~\eqref{eq:diversity_function} is monotone submodular.

Constructing a positive semidefinite kernel is well studied, and we can do it very flexibly.
For example, with any distance function $d: \mathcal{X} \times \mathcal{X} \rightarrow \mathbb{R}^{+}$,
$k(x, x') = \exp (-\beta \cdot d(x, x'))$
where $\beta \in \mathbb{R}^{+}$ is a scaling constant can be a positive semidefinite kernel~\cite{haasdonk2004learning}.
We show concrete examples and their interpretation in Appendix~\ref{appendix:div}.

\subsection{Computational Cost Reduction}
\label{subsec:costreduction}

Depending on the computational resource of the device we want to run our framework on,
reducing the computation is necessary, even though the performance is sacrificed.
We describe the possibilities for cost reduction below.

\paragraph{Determinant Calculation}
One of the heaviest computation in our framework is the determinant evaluation
in Eq.~\eqref{eq:diversity_function}, which is usually $O(n^3)$.
However, if we focus on the fact that $M$ is incrementally constructed,
that is, $M_{n+1}$ for the set $S \cup \{x_{n+1}\}$ always includes $M_n$
for the set $S$ as the leading principal submatrix, we can make this calculation $O(n^2)$
by elementary linear algebra (details in Appendix~\ref{appendix:det}).
Also, there exists a $O(n)$ approximation algorithm
for log determinants~\cite{dong2017scalable}.
When we can tolerate approximation errors, this can be a reasonable choice.

\paragraph{Dividing $K$}
As described above, one of the heaviest computation is the determinant calculation,
and it depends on the maximum batch size $K$.
Thus, one possible approach to decrease computation is
to decompose $K$ into $ \{K_1, K_2, \cdots K_n \}$ where $ \sum_i K_i = K$.
To simplify the problem, let $K_1 = K_2 = \cdots = K_n = K/n$.
Then, the computation cost for each $K_i$ is $O((K/n)^{3})$, and the total computation becomes $O(K^3/n^2)$, compared to the initial $O(K^3)$.
To take advantage of this relaxation, we should make an assumption such as
``no similar data appear in different divided substreams''.
Note that this is different from initially defining small $K$.
If $K$ is \textit{divided}, the model is trained after multiple batch selection (i.e., after selection of $n$ batches of size $K/n$),
while it is done after a single batch selection of $K$ samples in the default setting.

\paragraph{Memoization}
When we can use extra memory, \textit{memoization} is a promising approach to decrease computation.
The most effective element to use memoization on is the value of $k(\cdot, \cdot)$.
Although this is a standard technique, it is especially efficient when we use \SSPP
because it often calculates the same $k(\cdot, \cdot)$ on different sieves.

\section{Experiments}
\label{sec:experiments}

In this section, we evaluate our framework's effectiveness
and show the construction of concrete objective functions for specific tasks.
To show how broadly our approach can be applied,
we perform experiments with various datasets, tasks, and settings.
At first, we evaluate an image classification task using some toy datasets.
After that, we evaluate a more complex task, object detection, with a large practical dataset.
We also evaluate our framework on a real device (Raspberry Pi 4B).
If not specified otherwise, for all experiments with our method, only memoization is used among the computational cost reduction options
discussed in Section~\ref{subsec:costreduction}.
Our experiments on a server have been run on an Intel Xeon E5-2623 CPU with an Nvidia Tesla V100 GPU machine.

\subsection{Image Classification}

\begin{figure}[!t]
  \begin{minipage}[t]{\linewidth}
    \centering
    \includegraphics[keepaspectratio, width=.49\textwidth]{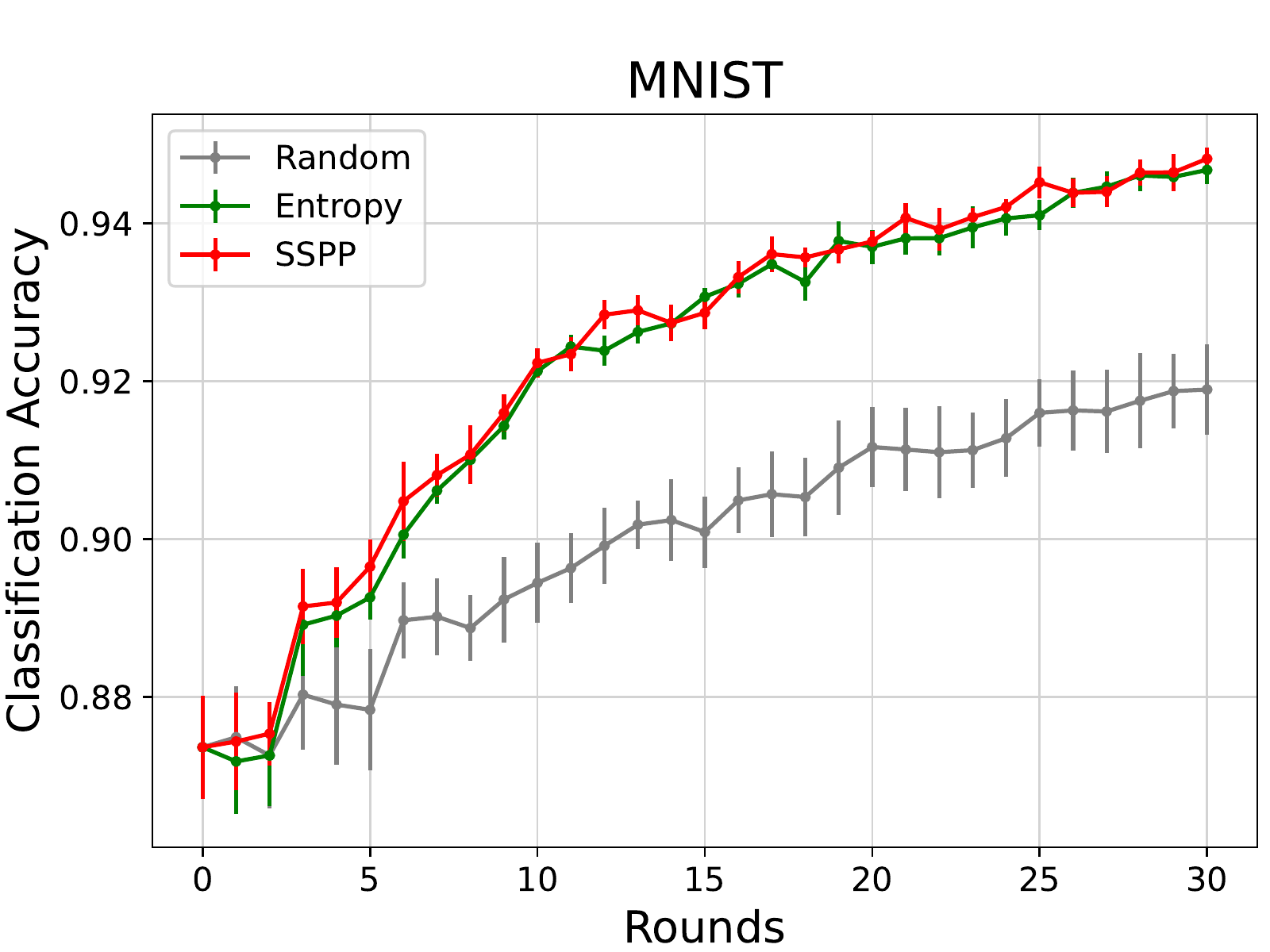}
    \includegraphics[keepaspectratio, width=.49\textwidth]{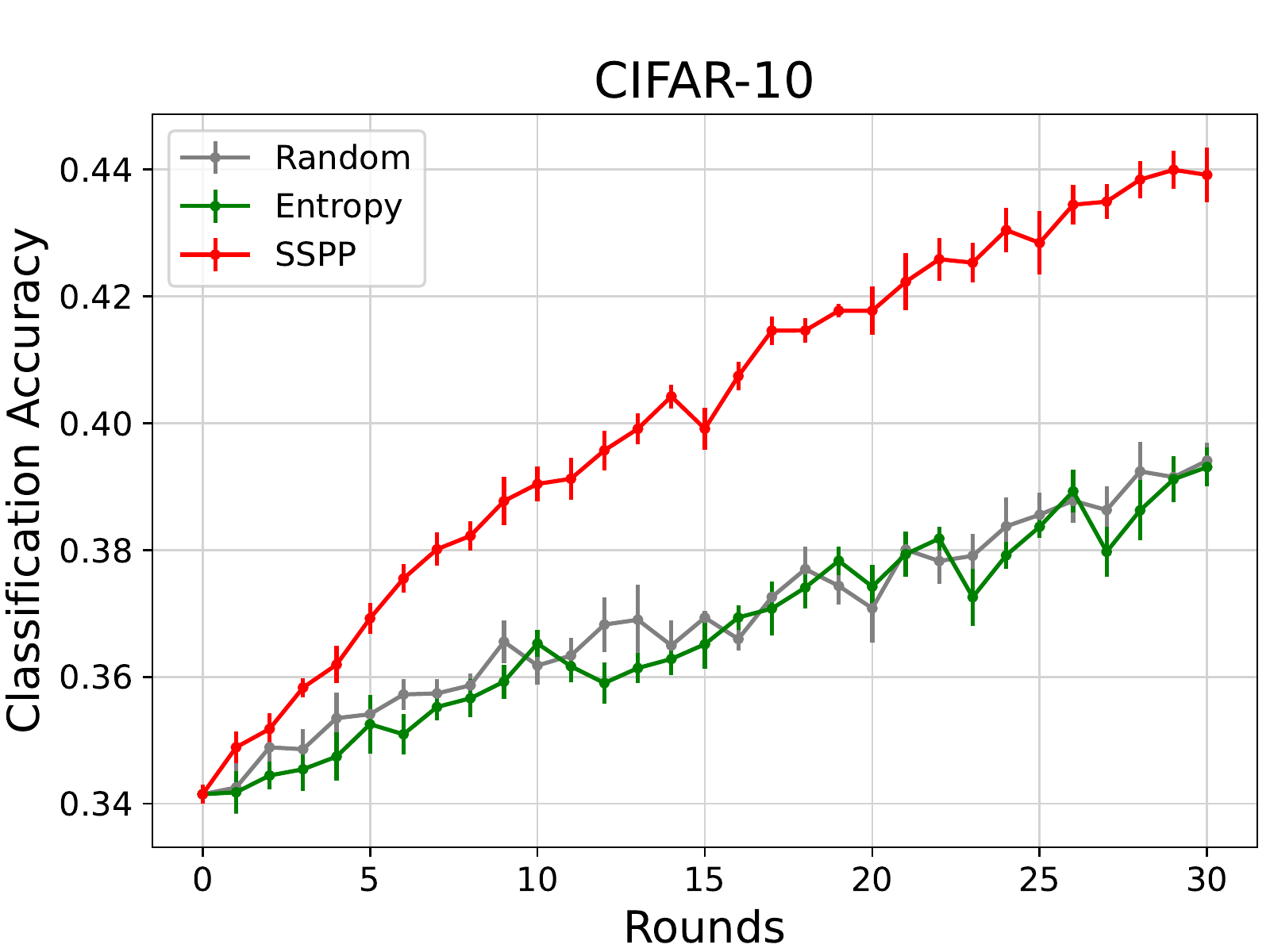}
    \subcaption{Cold-start scenario}
    \label{fig:classification_results_cs}
  \end{minipage}
  \begin{minipage}[t]{\linewidth}
    \centering
    \includegraphics[keepaspectratio, width=.49\textwidth]{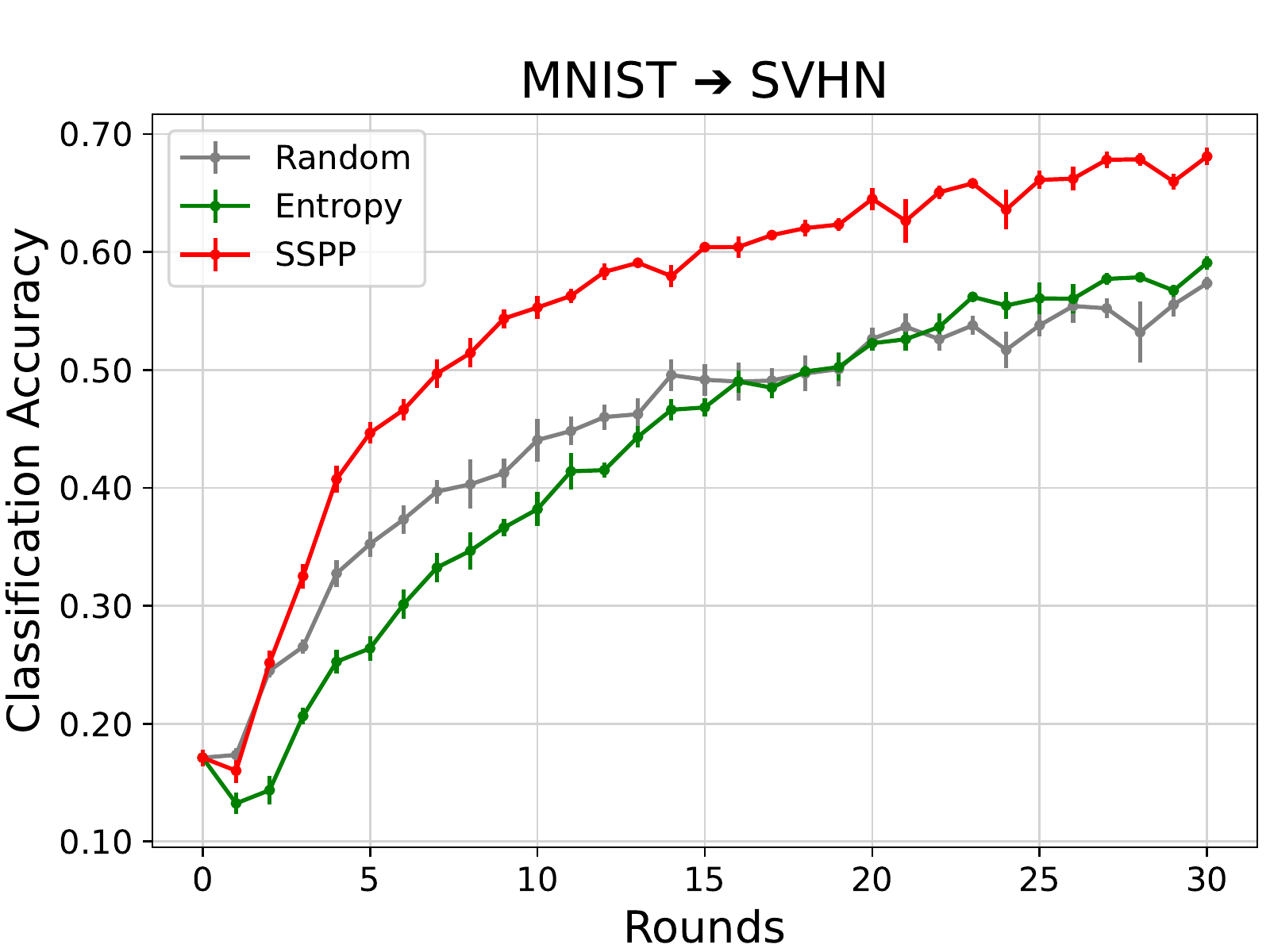}
    \includegraphics[keepaspectratio, width=.49\textwidth]{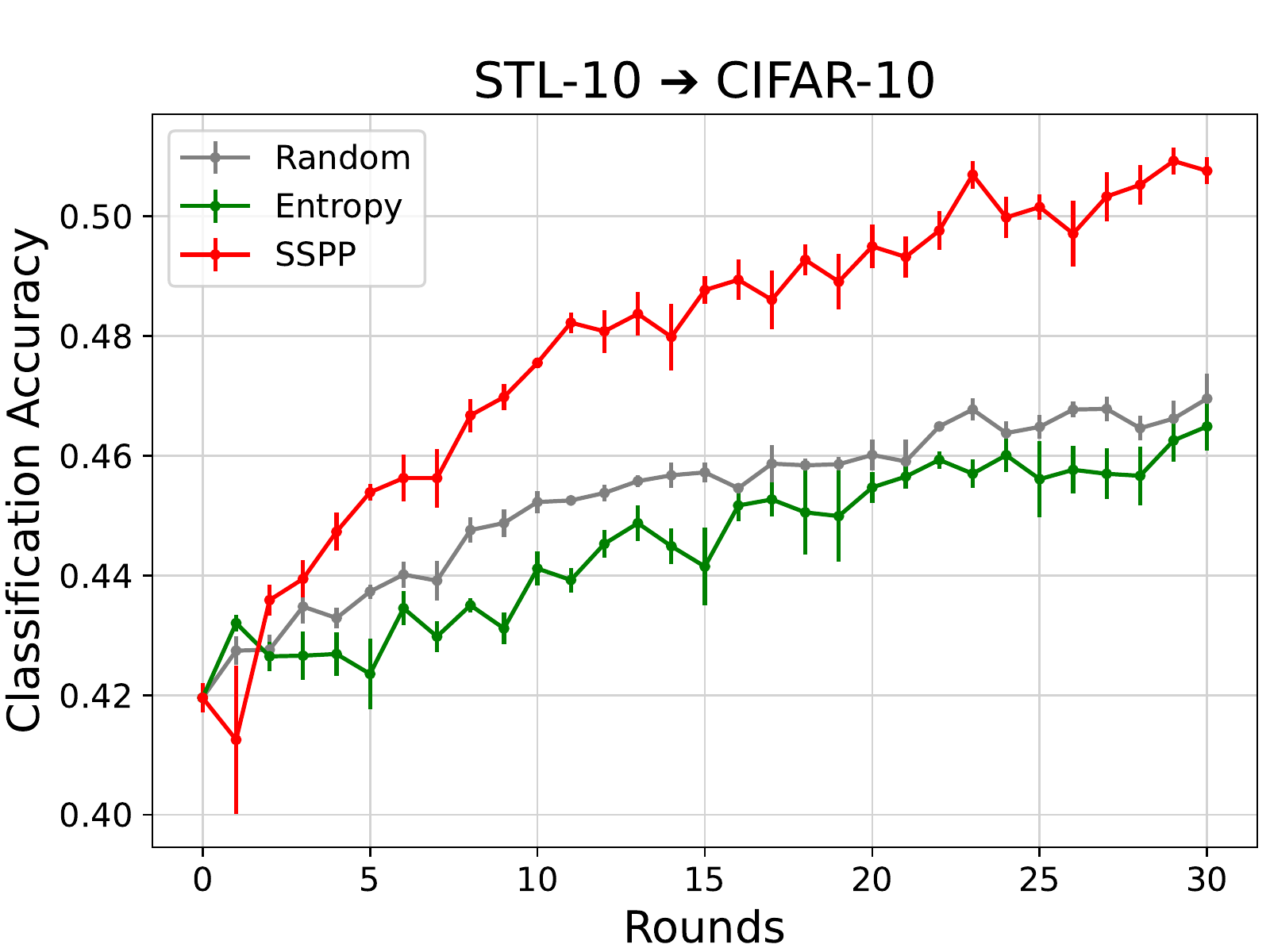}
    \subcaption{Domain adaptation scenario}
    \label{fig:classification_results_da}
  \end{minipage}
  \caption{
    Results of classification experiment
    in \subref{fig:classification_results_cs} the cold-start scenario
    and \subref{fig:classification_results_da} the domain adaptation scenario
    (mean accuracy vs. round number).
    Vertical bars show the std. err. across 5 repeats of the experiment
    with different random seeds.
  }
\end{figure}

For image classification task, we perform evaluations on both \textit{cold-start} and
\textit{domain adaptation} scenarios, described in Section~\ref{sec:problem_definition}.
In the experiments in this section, we are given $2048$ images sequentially (i.e., as a stream) for each round
and select up to $128$ from it.
The model is trained with all selected data in previous rounds.

\paragraph{Simulating data peculiarity}
As discussed in Section~\ref{sec:problem_definition}, we have to consider peculiarity of data.
To simulate class-imbalance, we divide the number of samples by $10$ for half of the categories,
similar to experiments in~\cite{aljundi2020identifying}.
Besides, all images are replicated four times,
blurred by Gaussian noise with a standard deviation of $0.01$ (i.e., 5 blurred similar images), and appear continuously on the stream
to simulate high-frequency camera capture as discussed in Section~\ref{sec:problem_definition}.
Also, to simulate the existence of many unuseful images,
we fill each stream with many ``non-object'' images ($1408$ images for our experiments).
Consequently, an additional class is added to the task, i.e., the former $C$-class classification
becomes a $(C+1)$-class classification, but instances of this added class are not included in the evaluation dataset.

\paragraph{Methods}
To solve the stream submodular optimization problem,
we use the known approximation algorithms \SSPP~\cite{kazemi2019submodular}
(which we denote as \texttt{SSPP}) with $\epsilon=0.1$.
As a comparison to our method, we consider an extension of a classic active learning method:
To the best of our knowledge, there are no existing methods for the problem we consider,
but some standard active learning methods can be adapted with trivial modifications.
For example, \textit{uncertainty sampling with entropy} \cite{settles2009active},
which is a one-by-one pool-based method, can be extended such that it
maximizes the sum of sample entropy in the batch (see detailed description in Appendix~\ref{appendix:algo_overview}).
To evaluate the importance of considering the problem-specific properties
discussed in Section \ref{sec:problem_definition}, we compare this method with ours.
Moreover, we also compare with random selection as a baseline.
We denote these methods as \texttt{Entropy} and \texttt{Random} respectively.

\paragraph{Objective Function}
As a measure of informativeness of each sample (i.e., $g$ in Eq.~\eqref{eq:informativeness_function}),
we use the entropy of softmax output~\cite{settles2009active}:
Let $F(x)_i$ be the i-th value of softmax output for the input $x$,
\begin{equation*}
  g(x) = \sum_i -F(x)_i \log F(x)_i.
\end{equation*}
To evaluate the diversity of a set with Eq.~\eqref{eq:diversity_function},
we define $k$ as polynomial kernel, which is positive definite~\cite{hofmann2008kernel}:
For a given pair of images $x, x' \in \mathcal{X}$, their similarity is evaluated as
$k(x, x') = \langle \hat{F}(x), \hat{F}(x') \rangle$
where $\hat{F}(x)$ is a middle layer output.

\subsubsection{Cold-start Scenario}
In this experiment, we use the popular datasets MNIST~\cite{lecun1998gradientbased},
Street View House Numbers (SVHN)~\cite{netzer2011reading},
and CIFAR-10~\cite{krizhevsky2009learning},
which consists of color images of objects from 10 classes.
As a cold-start setting, we set $|D_0| = 1000$ and training from scratch in each round,
i.e., after selecting a batch of up to $128$ images from the stream.
After the training, the model is evaluated on the \textit{test} data in each dataset.

The results are shown in Figure~\ref{fig:classification_results_cs} (SVHN result is shown in Appendix~\ref{appendix:more_exp}).
At first, we can see that our proposed framework with \texttt{SSPP} achieves the best performance
for all dataset.
Interestingly, only in MNIST experiments, the easiest task here,
\texttt{Entropy} also seems to work well.
In the other experiments, \texttt{Entropy} is as good as or even worse than \texttt{Random},
even though it uses the same informativeness criteria as \texttt{SSPP}.
These results show the importance of considering diversity
and suggest that its influence is stronger on more complex tasks.
The effect of considering diversity is also reflected in the unique number of selected images.
As the images in a stream are duplicated in this experiment,
the number of unique images is a good indicator to see how diverse were the batches selected.
Through all experiments, this number is \texttt{SSPP} $>$ \texttt{Random} $>$ \texttt{Entropy}
and the highest is three times the lowest.
We report the precise numbers in Appendix~\ref{appendix:more_exp} (Table~\ref{table:selected_samples}).

\subsubsection{Domain Adaptation Scenario}
For the domain adaptation scenario, we perform our evaluation on two pairs of datasets,
MNIST $\leftrightarrow$ SVHN and STL-10~\cite{coates2011analysis} $\rightarrow$ CIFAR-10~\footnote{
  Since STL-10 is small to construct a stream from,
  we do not consider CIFAR-10 $\rightarrow$ STL-10.
}.
We removed one class from both datasets in the latter setting
since there are only 9 classes in common.
As a domain adaptation scenario, the source dataset is available from the beginning.
At each round, we train a model after appending the initial (source) dataset with images from the other dataset (i.e., target dataset).
The evaluation is done on the \textit{test} split of the target dataset.

The results are shown in figure~\ref{fig:classification_results_da}
(SVHN $\rightarrow$ MNIST result is shown in Appendix~\ref{appendix:more_exp}).
Similar to the cold-start experiments, we observe that our approach with \texttt{SSPP} outperforms all others.
The tendency is the same as the results of the other setting,
it is a good indication that our framework performs well regardless of the scenario it is applied on.
The numbers about selected images are reported in Appendix~\ref{appendix:more_exp} (Table~\ref{table:selected_samples}).

\subsection{Object Detection}

Detection of objects in an image is a complex task requiring relatively heavy computation.
However, several models designed with a good balance between performance and latency exist,
coping with an edge-computing setup~\cite{tan2019mnasnet, zhang2017shufflenet, sandler2018mobilenetv2}.
In this domain-adaptation experiment, we use the approach Single-Shot Multibox Detector (SSD)~\cite{liu2016ssd} with the popular backbone Mobile Net v2~\cite{sandler2018mobilenetv2}.

We use two datasets in the experiments, the popular COCO (2017) dataset~\cite{lin2015microsoft} as the source dataset, which 
includes images with bounding box annotations of many categories, and BDD100K~\cite{yu2020bdd100k} as the target dataset,
which is a large-scale dataset of driving videos. Most frames in the BDD100K dataset videos are not
annotated with bounding box information, so we use the state-of-the-art detector EfficientDet~\cite{tan2020efficientdet} as an oracle for annotating
selected image samples.
The videos of BDD100K dataset are captured at 30 FPS, creating a stream with many redundant images.
In this setting, selecting a diversified batch of samples is very challenging, and it is well suited for 
evaluating the performance of our method.
We compare the same algorithms as in classification for solving the stream submodular
optimization problem, with parameter $\epsilon=0.01$ set for \texttt{SSPP}, if not specified otherwise.

\paragraph{Objective function}
The detection task involves both regression of bounding box parameters
and classification of object label, so ideally, we want a measure of the informativeness that copes with both aspects.
In the context of active learning with detection tasks,
several uncertainty functions designed for object detection are proposed~\cite{kao2019localizationaware}.
We use a combination of localization stability and classification uncertainty as the objective function
because those two do not rely on the architecture of the model. 
Keeping the notation of the paper, i.e., respectively $S_I(x)$ and $U_C(x)$:
\begin{equation*}
  g(x) = \lambda \cdot (1 - S_I(x)) + (1 - \lambda) \cdot U_C(x)
\end{equation*}
with $\lambda$ a tradeoff parameter between classification uncertainty and localization stability (fixed to $0.5$ in our experiments).
For evaluating the diversity of a set, we use the same kernel function $k$ as the image classification
experiments, which is task-independent.
The middle layer output $\hat{F}(x)$ used in this experiment is the concatenation of feature maps of the model, just before feeding the SSD extractor layers.

\subsubsection{Domain Adaptation Scenario}

\begin{figure}[!t]
  \centering
  \includegraphics[keepaspectratio, width=.29\textwidth]{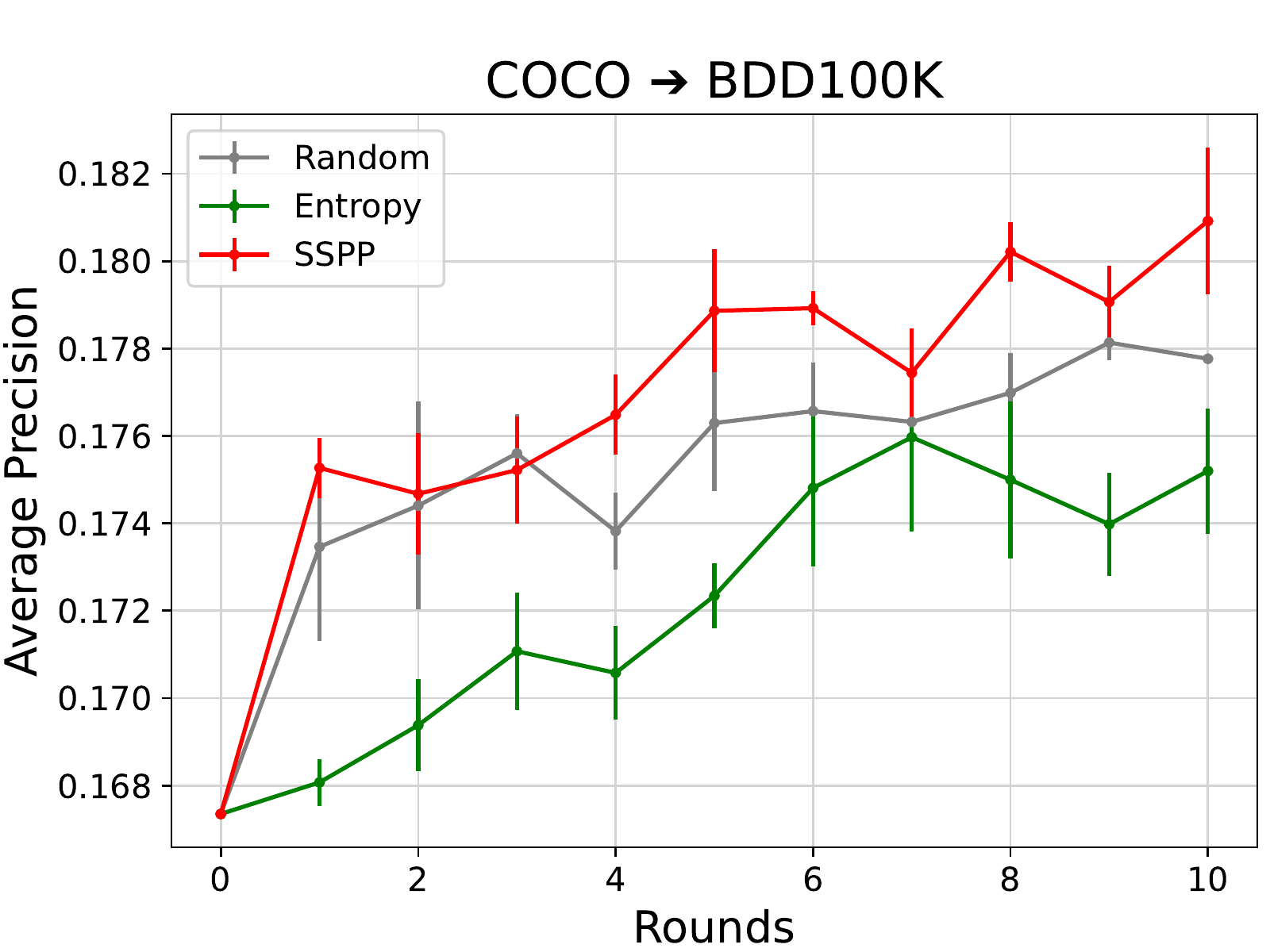}
  \caption{
    Results of domain adaptation experiment (average precision vs. round number). Vertical bars show the std. err. across 4 repeats of the experiment with different random seeds.
  }
  \label{fig:detection_results_da0}
\end{figure}
In this experiment, we initially train the detector on $D_0$, the train part of COCO dataset.
Then, for each round we select a batch of images from a stream composed of concatenated videos of BDD100k dataset
(the stream used in each round starts at the end of the previous one), with each round's stream consisting of $10000$ frames.
Since the class of objects in the detection task must exist in both source and target datasets,
we used the class \textit{bus}; instances of it are not omnipresent in street footages, so selecting samples containing instances is non-trivial.
The selected data $S_n$ is annotated by the oracle and appended to the previous dataset $D_n = D_{n-1} \cup S_n$.
After training, the new model performance is evaluated on the \textit{validation} split of BDD100K.

The results are shown in Figure~\ref{fig:detection_results_da0}.
As for classification tasks, \texttt{SSPP} outperforms the other methods, showing the
potential of this approach for complex tasks.
\texttt{Entropy} curves is however below the \texttt{Random} baseline.
The streams are composed of videos with images taken at high frequency, it is then not surprising that
\texttt{Entropy} approach performs poorly, as it lacks a mechanism taking the diversity into account.
We provide in Appendix~\ref{appendix:more_exp} both a comparison between \texttt{SSPP} with different values of $\epsilon$, and examples of batch selected by each algorithm.

\subsection{Real World Evaluation}

While our experiments above were conducted on a server for evaluation and reproducibility purposes,
our framework is designed to run on edge devices.
The two main issues for programs running on edge devices are the the memory footprint and the execution speed, so we perform a benchmark of the different methods we studied for both aspects.

\subsubsection{Execution Speed}
We compare the execution speed of the methods used in the previous experiments.
As the informativeness part of the objective function complexity depends on the model prediction,
for a fair comparison, we assert that the input images are the same for all measurements.
Thus, instead of capturing images from a camera, we load images stored beforehand on the device.
The source code for our framework used on the device is the same as the one used in server experiments above running on CPU,
but as the model inference is the bottleneck of computation, we optimized the inference computation using GPGPU~\footnote{
  We used the implementation publicly available at \url{https://github.com/Idein/py-videocore6} (GPL).
}.

For each method, we measure the elapsed time between the processing of two successive inputs and average it on $1000$ iterations.
Results are compiled in Table~\ref{table:rpi4speed}. The first line is representing inference speed when no selection is performed, for reference.
Firstly, we observe that the speed for all selection techniques are in the same order of magnitude, around 1 FPS.
\texttt{Entropy} uses a greedy approach, so besides evaluating the objective function,
there is close to no computation overhead.
The best performing \texttt{SSPP} is also the most computation-intensive, and because of the high number of sieves it uses,
both reducing $K$ or augmenting $\epsilon$ have a significant impact on the speed. The $\epsilon$ parameter directly influences
the number of sieves considered during the computation in \texttt{SSPP} and is a tradeoff parameter between speed and accuracy.
At fixed $\epsilon$, each sieve's size depends on $K$, and so does the size of the matrix to compute the determinant of, for the evaluation diversity.
This confirms the potential of \textit{Dividing $K$} discussed in Section~\ref{sec:approach}.
To prove the soundness of our approach, we also show at the second line the elapsed time when performing inference and data sending over HTTP performed for each sample.
It corresponds to the case discussed in Section~\ref{sec:introduction}, where the selection of samples is done on a server. The computation overhead due to sending data
is too heavy to allow a practical speed, with the average elapsed time an order of magnitude higher compared to the other.
This confirms the necessity of screening samples on the device and only sending selected samples.
\begin{table}[!t]
  \centering
  \caption{Active learning framework speed measured on Raspberry 4B. Results expressed in elapsed time (seconds), with the mean and std. err. reported over $1000$ iterations. (The lower, the better)}
  \small
  \begin{tabular}{ccc}
    \hline
    {} & \multicolumn{2}{c}{Batch size} \\
    {} & $K=16$ & $K=32$  \\ \hline
    (inference only) & (0.19{\tiny $\pm$0.00}) & (0.19{\tiny $\pm$0.00})  \\
    (Sending all data) & 3.11{\tiny $\pm$0.03} & 3.11{\tiny $\pm$0.03}  \\
    \texttt{Entropy} & 0.55{\tiny $\pm$0.02} & 0.55{\tiny $\pm$0.02}  \\
    \texttt{SSPP} ($\epsilon = 0.1$) & 0.57{\tiny $\pm$0.02} & 0.61{\tiny $\pm$0.02}  \\
    \texttt{SSPP} ($\epsilon = 0.05$) & 0.59{\tiny $\pm$0.02} & 0.66{\tiny $\pm$0.02}  \\
    \texttt{SSPP} ($\epsilon = 0.01$) & 0.78{\tiny $\pm$0.03} & 1.23{\tiny $\pm$0.05}  \\

  \hline
  \end{tabular}
  \label{table:rpi4speed}
\end{table}

\subsubsection{Memory Usage}
While Memory Swapping virtually increases the available memory, performing too much read/write on the storage device (e.g. SD cards) can
harm its longevity and/or degrade execution speed.
Among our studied methods, \texttt{SSPP} has the highest space complexity, so we
measured its memory footprint during the experiments, with and without memoization discussed in Section~\ref{sec:approach}.
While memoization is mainly used to reduce execution speed at the cost of more memory use, when enabling memoization
we also introduced a look-up table to store only once images that are present in multiple sieves, allowing
to reduce the memory footprint of the studied methods.
With $K = 32$ and $\epsilon = 0.01$, the average memory usage obtained were $3626$ MB and $5316$ MB.
The first observation is that efficient caching of samples allows to significantly reduce the memory footprint (by more than $30\%$).
Secondly, even for the most memory-heavy setting of Table~\ref{table:rpi4speed}, the mean consumption stayed under $4$ GB,
meaning that all settings allows an operation without relying on memory swapping for most edge-devices.

\section{Conclusion}

We formalized active learning for DNNs on edge devices.
Based on the considerations about the nature of the problem,
a solution method is required to be 1) stream-based and 2) in batch-mode.
We proposed a general framework to solve this problem as a stream submodular maximization problem,
which takes both the informativeness of sole data and the diverseness of a set of data into account.
We evaluated our approach on classification and object detection tasks
with several datasets on servers and real-world edge devices.
Empirical results showed that our approach outperforms other methods and can run even on low computational resources devices.

{\small
\bibliographystyle{ieee_fullname}

\begin{thebibliography}{10}\itemsep=-1pt

\bibitem{aljundi2020identifying}
Rahaf Aljundi, Nikolay Chumerin, and Daniel~Olmeda Reino.
\newblock Identifying {{Wrongly Predicted Samples}}: {{A Method}} for {{Active
  Learning}}.
\newblock {\em arXiv:2010.06890}, 2020.

\bibitem{amin2020understanding}
Kareem Amin, Corinna Cortes, Giulia DeSalvo, and Afshin Rostamizadeh.
\newblock Understanding the {{Effects}} of {{Batching}} in {{Online Active
  Learning}}.
\newblock In {\em International {{Conference}} on {{Artificial Intelligence}}
  and {{Statistics}}}, pages 3482--3492. {PMLR}, 2020.

\bibitem{ash2020deep}
Jordan~T. Ash, Chicheng Zhang, Akshay Krishnamurthy, John Langford, and Alekh
  Agarwal.
\newblock Deep {{Batch Active Learning}} by {{Diverse}}, {{Uncertain Gradient
  Lower Bounds}}.
\newblock In {\em International {{Conference}} on {{Learning
  Representations}}}, 2020.

\bibitem{badanidiyuru2014streaming}
Ashwinkumar Badanidiyuru, Baharan Mirzasoleiman, Amin Karbasi, and Andreas
  Krause.
\newblock Streaming submodular maximization: Massive data summarization on the
  fly.
\newblock In {\em Proceedings of the 20th {{ACM SIGKDD}} International
  Conference on {{Knowledge}} Discovery and Data Mining - {{KDD}} '14}, pages
  671--680. {ACM Press}, 2014.

\bibitem{buschjager2020very}
Sebastian Buschj{\"a}ger, Philipp-Jan Honysz, and Katharina Morik.
\newblock Very {{Fast Streaming Submodular Function Maximization}}.
\newblock {\em arXiv:2010.10059}, 2020.

\bibitem{coates2011analysis}
Adam Coates, Honglak Lee, and Andrew Ng.
\newblock An {{Analysis}} of {{Single}}-{{Layer Networks}} in {{Unsupervised
  Feature Learning}}.
\newblock In {\em Proceedings of the {{Fourteenth International Conference}} on
  {{Artificial Intelligence}} and {{Statistics}}}, pages 215--223. {JMLR
  Workshop and Conference Proceedings}, 2011.

\bibitem{dong2017scalable}
Kun Dong, David Eriksson, Hannes Nickisch, David Bindel, and Andrew~G. Wilson.
\newblock Scalable {{Log Determinants}} for {{Gaussian Process Kernel
  Learning}}.
\newblock {\em Advances in Neural Information Processing Systems},
  30:6327--6337, 2017.

\bibitem{freytag2014selecting}
Alexander Freytag, Erik Rodner, and Joachim Denzler.
\newblock Selecting {{Influential Examples}}: {{Active Learning}} with
  {{Expected Model Output Changes}}.
\newblock In {\em European {{Conference}} on {{Computer Vision}}}, pages
  562--577, 2014.

\bibitem{gal2017deep}
Yarin Gal, Riashat Islam, and Zoubin Ghahramani.
\newblock Deep {{Bayesian Active Learning}} with {{Image Data}}.
\newblock In {\em International {{Conference}} on {{Machine Learning}}}, pages
  1183--1192. {PMLR}, 2017.

\bibitem{ganin2016domainadversarial}
Yaroslav Ganin, Evgeniya Ustinova, Hana Ajakan, Pascal Germain, Hugo
  Larochelle, Fran{\c{c}}ois Laviolette, Mario March, and Victor Lempitsky.
\newblock Domain-adversarial training of neural networks.
\newblock {\em Journal of Machine Learning Research}, 17(59):1--35, 2016.

\bibitem{girshick2014rich}
Ross Girshick, Jeff Donahue, Trevor Darrell, and Jitendra Malik.
\newblock Rich {{Feature Hierarchies}} for {{Accurate Object Detection}} and
  {{Semantic Segmentation}}.
\newblock In {\em Proceedings of the {{IEEE Conference}} on {{Computer Vision}}
  and {{Pattern Recognition}}}, pages 580--587, 2014.

\bibitem{haasdonk2004learning}
Bernard Haasdonk and Claus Bahlmann.
\newblock Learning with {{Distance Substitution Kernels}}.
\newblock In {\em Pattern {{Recognition}}}, Lecture {{Notes}} in {{Computer
  Science}}, pages 220--227. {Springer}, 2004.

\bibitem{hofmann2008kernel}
Thomas Hofmann, Bernhard Sch{\"o}lkopf, and Alexander~J. Smola.
\newblock Kernel methods in machine learning.
\newblock {\em Annals of Statistics}, 36(3):1171--1220, 2008.

\bibitem{hoi2006batch}
Steven C.~H. Hoi, Rong Jin, Jianke Zhu, and Michael~R. Lyu.
\newblock Batch mode active learning and its application to medical image
  classification.
\newblock In {\em Proceedings of the 23rd International Conference on
  {{Machine}} Learning}, {{ICML}} '06, pages 417--424. {Association for
  Computing Machinery}, 2006.

\bibitem{howard2019searching}
Andrew Howard, Mark Sandler, Grace Chu, Liang-Chieh Chen, Bo Chen, Mingxing
  Tan, Weijun Wang, Yukun Zhu, Ruoming Pang, Vijay Vasudevan, Quoc~V. Le, and
  Hartwig Adam.
\newblock Searching for {{MobileNetV3}}.
\newblock In {\em Proceedings of the {{IEEE}}/{{CVF International Conference}}
  on {{Computer Vision}}}, pages 1314--1324, 2019.

\bibitem{joshi2009multiclass}
A.~J. Joshi, F. Porikli, and N. Papanikolopoulos.
\newblock Multi-class active learning for image classification.
\newblock In {\em 2009 {{IEEE Conference}} on {{Computer Vision}} and {{Pattern
  Recognition}}}, pages 2372--2379, 2009.

\bibitem{kao2019localizationaware}
Chieh-Chi Kao, Teng-Yok Lee, Pradeep Sen, and Ming-Yu Liu.
\newblock Localization-{{Aware Active Learning}} for {{Object Detection}}.
\newblock In {\em Computer {{Vision}} \textendash{} {{ACCV}} 2018}, Lecture
  {{Notes}} in {{Computer Science}}, pages 506--522. {Springer International
  Publishing}, 2019.

\bibitem{kaushal2019learning}
V. Kaushal, R. Iyer, S. Kothawade, R. Mahadev, K. Doctor, and G. Ramakrishnan.
\newblock Learning {{From Less Data}}: {{A Unified Data Subset Selection}} and
  {{Active Learning Framework}} for {{Computer Vision}}.
\newblock In {\em 2019 {{IEEE Winter Conference}} on {{Applications}} of
  {{Computer Vision}}}, pages 1289--1299, 2019.

\bibitem{kazemi2019submodular}
Ehsan Kazemi, Marko Mitrovic, Morteza Zadimoghaddam, Silvio Lattanzi, and Amin
  Karbasi.
\newblock Submodular {{Streaming}} in {{All Its Glory}}: {{Tight
  Approximation}}, {{Minimum Memory}} and {{Low Adaptive Complexity}}.
\newblock In {\em International {{Conference}} on {{Machine Learning}}}, pages
  3311--3320. {PMLR}, 2019.

\bibitem{kingma2015adam}
Diederik~P Kingma and Jimmy Ba.
\newblock Adam: A method for stochastic optimization.
\newblock In {\em International Conference on Learning Representations (ICLR)},
  2015.

\bibitem{kirsch2019batchbald}
Andreas Kirsch, Joost van Amersfoort, and Yarin Gal.
\newblock {{BatchBALD}}: {{Efficient}} and {{Diverse Batch Acquisition}} for
  {{Deep Bayesian Active Learning}}.
\newblock In {\em Advances in {{Neural Information Processing Systems}}}, pages
  7026--7037, 2019.

\bibitem{krause2012submodular}
Andreas Krause and Daniel Golovin.
\newblock Submodular {{Function Maximization}}.
\newblock In {\em Tractability}, pages 71--104. {Cambridge University Press},
  2012.

\bibitem{krizhevsky2009learning}
Alex Krizhevsky and Geoffrey Hinton.
\newblock Learning {{Multiple Layers}} of {{Features}} from {{Tiny Images}}.
\newblock Technical report, 2009.

\bibitem{kulesza2012determinantal}
Alex Kulesza and Ben Taskar.
\newblock Determinantal point processes for machine learning.
\newblock {\em Foundations and Trends\textregistered{} in Machine Learning},
  5(2-3):123--286, 2012.

\bibitem{lawrence2003fast}
Neil Lawrence, Matthias Seeger, and Ralf Herbrich.
\newblock Fast sparse gaussian process methods: {{The}} informative vector
  machine.
\newblock In {\em Advances in Neural Information Processing Systems},
  volume~15, pages 625--632. {MIT Press}, 2003.

\bibitem{lecun1998gradientbased}
Yann LeCun, L{\'e}on Bottou, Yoshua Bengio, and Patrick Haffner.
\newblock Gradient-based learning applied to document recognition.
\newblock {\em Proceedings of the IEEE}, 86(11):2278--2324, 1998.

\bibitem{lin2015microsoft}
Tsung-Yi Lin, Michael Maire, Serge Belongie, Lubomir Bourdev, Ross Girshick,
  James Hays, Pietro Perona, Deva Ramanan, C.~Lawrence Zitnick, and Piotr
  Doll^^c3^^a1r.
\newblock Microsoft coco: Common objects in context, 2015.

\bibitem{liu2016ssd}
Wei Liu, Dragomir Anguelov, Dumitru Erhan, Christian Szegedy, Scott Reed,
  Cheng-Yang Fu, and Alexander~C. Berg.
\newblock {{SSD}}: {{Single Shot MultiBox Detector}}.
\newblock In {\em European {{Conference}} on {{Computer Vision}}}, pages
  21--37, 2016.

\bibitem{long2015learning}
Mingsheng Long, Yue Cao, Jianmin Wang, and Michael Jordan.
\newblock Learning transferable features with deep adaptation networks.
\newblock In {\em Proceedings of the {{IEEE International Conference}} on
  {{Computer Vision}}}, pages 97--105. PMLR, 2015.

\bibitem{nemhauser1978analysis}
George~L. Nemhauser, Laurence~A. Wolsey, and Marshall~L. Fisher.
\newblock An analysis of approximations for maximizing submodular set
  functions\textemdash{{I}}.
\newblock {\em Mathematical Programming}, 14(1):265--294, 1978.

\bibitem{netzer2011reading}
Yuval Netzer, Tao Wang, Adam Coates, Alessandro Bissacco, Bo Wu, and Andrew Ng.
\newblock Reading {{Digits}} in {{Natural Images}} with {{Unsupervised Feature
  Learning}}.
\newblock {\em NIPS}, 2011.

\bibitem{DBLP:journals/corr/abs-1912-01703}
Adam Paszke, Sam Gross, Francisco Massa, Adam Lerer, James Bradbury, Gregory
  Chanan, Trevor Killeen, Zeming Lin, Natalia Gimelshein, Luca Antiga, Alban
  Desmaison, Andreas K{\"{o}}pf, Edward Yang, Zach DeVito, Martin Raison,
  Alykhan Tejani, Sasank Chilamkurthy, Benoit Steiner, Lu Fang, Junjie Bai, and
  Soumith Chintala.
\newblock Pytorch: An imperative style, high-performance deep learning library.
\newblock {\em arXiv:1912.01703}, 2019.

\bibitem{roy2001optimal}
Nicholas Roy and Andrew McCallum.
\newblock Toward {{Optimal Active Learning}} through {{Sampling Estimation}} of
  {{Error Reduction}}.
\newblock In {\em International {{Conference}} on {{Machine Learning}}},
  {{ICML}} '01, pages 441--448. {Morgan Kaufmann Publishers Inc.}, 2001.

\bibitem{sandler2018mobilenetv2}
Mark Sandler, Andrew Howard, Menglong Zhu, Andrey Zhmoginov, and Liang-Chieh
  Chen.
\newblock {{MobileNetV2}}: {{Inverted Residuals}} and {{Linear Bottlenecks}}.
\newblock In {\em Proceedings of the {{IEEE Conference}} on {{Computer Vision}}
  and {{Pattern Recognition}}}, pages 4510--4520, 2018.

\bibitem{sener2018active}
Ozan Sener and Silvio Savarese.
\newblock Active {{Learning}} for {{Convolutional Neural Networks}}: {{A
  Core}}-{{Set Approach}}.
\newblock In {\em International {{Conference}} on {{Learning
  Representations}}}, 2018.

\bibitem{settles2009active}
Burr Settles.
\newblock Active {{Learning Literature Survey}}.
\newblock Technical {{Report}}, {University of Wisconsin-Madison Department of
  Computer Sciences}, 2009.

\bibitem{settles2008multipleinstance}
Burr Settles, Mark Craven, and Soumya Ray.
\newblock Multiple-instance active learning.
\newblock In {\em Advances in Neural Information Processing Systems},
  volume~20, pages 1289--1296. {Curran Associates, Inc.}, 2008.

\bibitem{sharma2015greedy}
Dravyansh Sharma, Ashish Kapoor, and Amit Deshpande.
\newblock On {{Greedy Maximization}} of {{Entropy}}.
\newblock In {\em International {{Conference}} on {{Machine Learning}}}, pages
  1330--1338. {PMLR}, 2015.

\bibitem{sun2017revisiting}
Chen Sun, Abhinav Shrivastava, Saurabh Singh, and Abhinav Gupta.
\newblock Revisiting {{Unreasonable Effectiveness}} of {{Data}} in {{Deep
  Learning Era}}.
\newblock In {\em Proceedings of the {{IEEE International Conference}} on
  {{Computer Vision}}}, pages 843--852, 2017.

\bibitem{pmlr-v28-sutskever13}
Ilya Sutskever, James Martens, George Dahl, and Geoffrey Hinton.
\newblock On the importance of initialization and momentum in deep learning.
\newblock In {\em International {{Conference}} on {{Machine Learning}}}, pages
  1139--1147. PMLR, 2013.

\bibitem{tan2019mnasnet}
Mingxing Tan, Bo Chen, Ruoming Pang, Vijay Vasudevan, Mark Sandler, Andrew
  Howard, and Quoc~V. Le.
\newblock {{MnasNet}}: {{Platform}}-{{Aware Neural Architecture Search}} for
  {{Mobile}}.
\newblock In {\em Proceedings of the {{IEEE}}/{{CVF Conference}} on {{Computer
  Vision}} and {{Pattern Recognition}}}, pages 2820--2828, 2019.

\bibitem{tan2019efficientnet}
Mingxing Tan and Quoc Le.
\newblock {{EfficientNet}}: {{Rethinking Model Scaling}} for {{Convolutional
  Neural Networks}}.
\newblock In {\em International {{Conference}} on {{Machine Learning}}}, pages
  6105--6114. {PMLR}, 2019.

\bibitem{tan2020efficientdet}
Mingxing Tan, Ruoming Pang, and Quoc~V. Le.
\newblock Efficientdet: Scalable and efficient object detection, 2020.

\bibitem{wang2021neural}
Zhilei Wang, Pranjal Awasthi, Christoph Dann, Ayush Sekhari, and Claudio
  Gentile.
\newblock Neural {{Active Learning}} with {{Performance Guarantees}}.
\newblock In {\em Advances in {{Neural Information Processing Systems}}},
  volume~34, pages 7510--7521. {Curran Associates, Inc.}, 2021.

\bibitem{wei2015submodularity}
Kai Wei, Rishabh Iyer, and Jeff Bilmes.
\newblock Submodularity in {{Data Subset Selection}} and {{Active Learning}}.
\newblock In {\em International {{Conference}} on {{Machine Learning}}}, pages
  1954--1963. {PMLR}, 2015.

\bibitem{yu2020bdd100k}
Fisher Yu, Haofeng Chen, Xin Wang, Wenqi Xian, Yingying Chen, Fangchen Liu,
  Vashisht Madhavan, and Trevor Darrell.
\newblock {{BDD100K}}: {{A Diverse Driving Dataset}} for {{Heterogeneous
  Multitask Learning}}.
\newblock In {\em Proceedings of the {{IEEE}}/{{CVF Conference}} on {{Computer
  Vision}} and {{Pattern Recognition}}}, pages 2636--2645, 2020.

\bibitem{zhang2017shufflenet}
Xiangyu Zhang, Xinyu Zhou, Mengxiao Lin, and Jian Sun.
\newblock Shufflenet: An extremely efficient convolutional neural network for
  mobile devices, 2017.

\end{thebibliography}

}

\newpage
\appendix

\section{About Algorithms}
\label{appendix:algo_overview}

\subsection{Stream submodular maximization algorithms}
As discussed in Section~\ref{sec:related_work}, three algorithms were proposed
to solve submodular maximization problems in a stream setting: \SieveStreaming, \SSPP and \ThreeSieves.
The base idea behind these three algorithms is to estimate the required marginal gain
for a current set to achieve $(1/2 - \epsilon)$ approximation.
In \SieveStreaming and \SSPP, the algorithm simultaneously holds several sets (called \textit{sieves}) and
their corresponding required marginal gain as a threshold.
When a new sample comes, each \textit{sieve} evaluates if
the marginal gain by adding the new sample is higher or lower than its threshold,
then respectively appends or discards it.
At the end of the data stream, the algorithm returns the \textit{sieve} with the highest set value as a solution.
In \ThreeSieves, instead of holding multiple thresholds,
the algorithm keeps only the highest threshold and gradually decreases it by statistical analysis.

\subsection{Extended simple AL algorithms}
Active learning methods from previous work designed for
non-batch setting often use an uncertainty function that
evaluates the informativeness of a single data sample.
It can be extended to the batch setting by taking the sum
of the contributions of samples in a set of data. This is a special
case of our framework without diversity, i.e., $\lambda=0$ in Eq.~\eqref{eq:objective_function}.
Note that in this case, Eq.~\eqref{eq:objective_function} is modular (linear),
and we can get the optimal solution by a greedy approach.

\subsection{Related methods}

We show a comparison of related methods in Table~\ref{table:related_al_methods}.
It emphasizes that we cannot apply existing methods to the problem we consider.

\begin{table}[!h]
  \caption{Related AL methods.}
  \label{table:related_al_methods}
  \small
    \centering
    \begin{tabular}{cccc}
      \hline
      {}                          & stream-based & batch-mode   \\ \hline
      \cite{wang2021neural}      & $\checkmark$ &              \\
      \cite{sener2018active}     &              & $\checkmark$ \\
      \cite{ash2020deep}         &              & $\checkmark$ \\
      Ours                        & $\checkmark$ & $\checkmark$ \\ \hline
    \end{tabular}
\end{table}

\section{Efficient Determinant Calculation}
\label{appendix:det}

By construction, a similarity matrix $M_{n+1}$ for $n \geq 1$ is described as follow:
\begin{equation*}
  M_{n+1} = \begin{pmatrix} M_n & v  \\ v^T & a \end{pmatrix}
\end{equation*}
\noindent
where $M_n \in \mathbb{R}^{n \times n}, v \in \mathbb{R}^n, a \in \mathbb{R}$.
Then, we can calculate determinant using $\det(M_n)$ as follow:
\begin{equation}
  \label{incremented_M}
  \det(M_{n+1}) = (a - v^T M_n^{-1} v) \det(M_n)
\end{equation}
\noindent
This is $O(n^2)$ when $\det(M_n)$ and $M_{n}^{-1}$ is given.
While calculating an inverse matrix is generally computation heavy,
it can be efficiently calculated when a matrix is defined as Eq.~\eqref{incremented_M}:
\begin{equation}
  \label{inverse_M}
  M_{n+1}^{-1} = \begin{pmatrix}
  M_n^{-1} + \frac{1}{s} M_n^{-1} v v^T M_n^{-1} & - \frac{1}{s} M_n^{-1} v \\
  - \frac{1}{s} v^T M_n^{-1} & \frac{1}{s}
  \end{pmatrix}
\end{equation}

\noindent
where $s = a - v^T M_n^{-1} v$.
Note that a similarity matrix is always symmetric, thus 
$v^T M_n^{-1} = (M_n^{-1} v)^T$
is always held.
Using this property, Eq.~\eqref{inverse_M} is also efficiently calculated.

\begin{figure}[!t]
  \vskip 0.15in
  \begin{minipage}[t]{\linewidth}
    \centering
    \includegraphics[keepaspectratio, width=.9\textwidth]{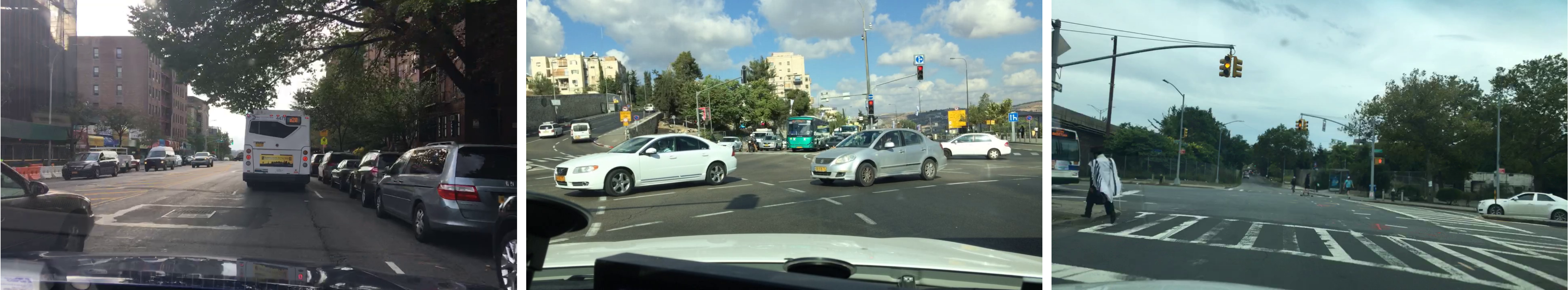}
    \subcaption{Imbalanced}
    \label{fig:peculiarity_imbalanced}
  \end{minipage}
  \vskip 0.15in
  \begin{minipage}[t]{\linewidth}
    \centering
    \includegraphics[keepaspectratio, width=.9\textwidth]{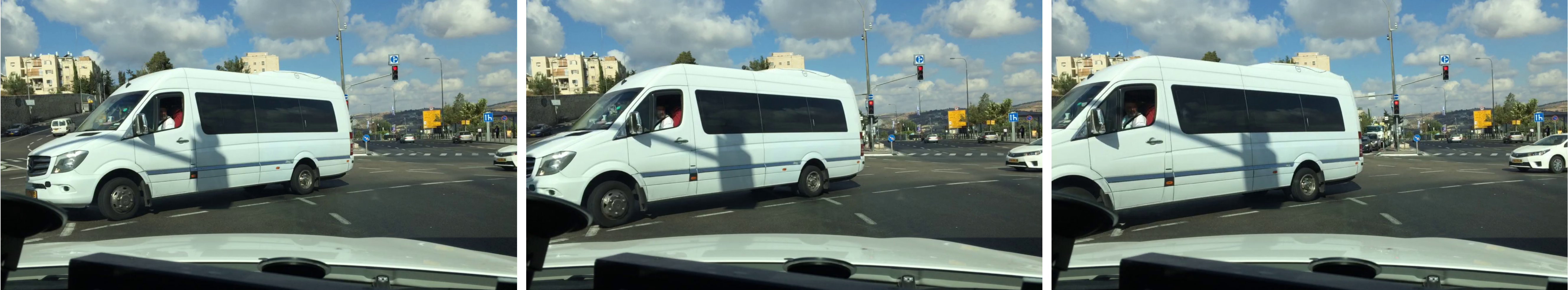}
    \subcaption{Duplicated}
    \label{fig:peculiarity_duplicated}
  \end{minipage}
  \vskip 0.15in
  \begin{minipage}[t]{\linewidth}
    \centering
    \includegraphics[keepaspectratio, width=.9\textwidth]{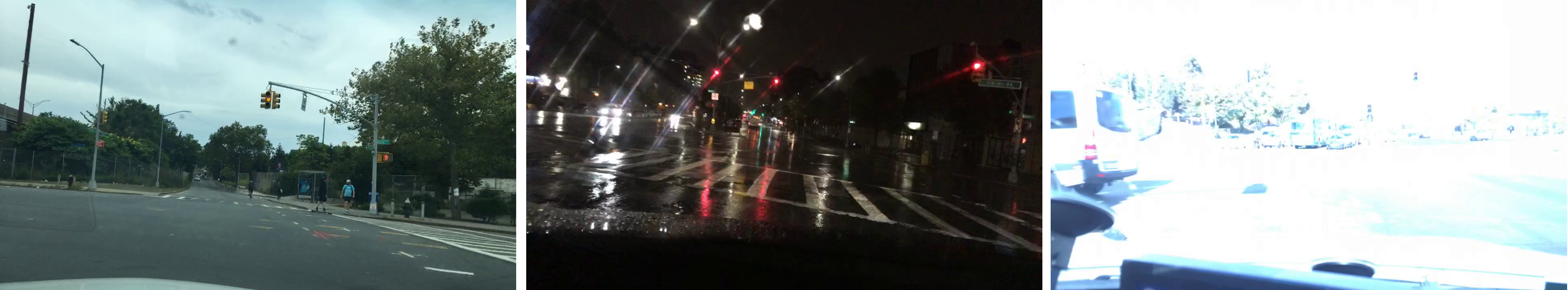}
    \subcaption{Unuseful}
    \label{fig:peculiarity_unuseful}
  \end{minipage}
  \caption{
    Examples of the peculiarity of data from BDD100K dataset.
    \subref{fig:peculiarity_imbalanced} Imbalanced: while there are a lot of ``Car'' class objects, ``Person'' class object is little represented.
    \subref{fig:peculiarity_duplicated} Duplicated: If the frequency of data capture is high, very similar images will appear.
    \subref{fig:peculiarity_unuseful} Unuseful: There are a lot of images which contain no objects, or failed to be captured correctly.
  }
  \label{fig:data_peculiarity}
\end{figure}

\section{Diversity Construction Details}
\label{appendix:div}

As discussed in Section~\ref{sec:approach},
we have to define a kernel function $k$ to evaluate the diversity of a set.
There are many possible choices for $k$ to measure the similarity between a pair of images $x, x' \in \mathcal{X}$.
For example,
\begin{align*}
  k_1(x, x') &= \langle \hat{F}(x), \hat{F}(x') \rangle \\
  k_2(x, x') &= \exp(-\beta \cdot ||x-x'||_1) \\
  k_3(x, x') &= \exp(-\beta \cdot ||\hat{F}(x)-\hat{F}(x')||_2) \\
  k_4(x, x') &= \exp(-\beta \cdot \text{JSD}(F(x) || F(x')))
\end{align*}
where $\hat{F}(x)$ is a middle layer output, $\beta$ is a scaling parameter and $\text{JSD}$ is
short-hand for Jensen-Shannon divergence.
Here, $k_1$ is known as a polynomial kernel and $k_2, k_3, k_4$ are known as an RBF kernel,
and both kernels are positive definite~\cite{hofmann2008kernel}.
This selection can be interpreted as choosing which space we focus on, i.e.,
consider the similarity in the image space, feature vector space, etc.

\section{More Implementation Details}
\label{appendix:impl_details}
Through all our experiments, we used the machine learning framework PyTorch~\cite{DBLP:journals/corr/abs-1912-01703}.

\subsection{Image Classification}
For the classification task, we use a two-layer convolutional neural network as a classifier.
We trained the model by Adam~\cite{kingma2015adam} optimizer with $12000$ iterations, batch size $16$.
In the beginning, it is randomly initialized and trained with an initial dataset $D_0$.
Then, for each round, we are given $2048$ images sequentially (i.e., as a stream) and
select up to $128$ from it.
These images are appended to the current dataset ($D_{n-1}$ at round $n$).
In each round, we trained the model from scratch.
As a simulation of manual annotation, the labels for each image are given after selection
so that the labels are unknown during selection.
In domain adaptation experiments, we equally sample from source domain images and
target domain images  to form training batches, since they are highly imbalanced, especially in early rounds.
For the ``non-object'' image, we used a monochrome image obtained by taking the mean color (i.e., the means of each RGB channel) of the next real image in the stream.
The elapsed time per round was a few minutes in each experiment.

\subsection{Object Detection}
For the detection task, we based our model on a PyTorch implementation of SSD~\footnote{
  \url{https://github.com/lufficc/SSD} (MIT).
}.
The training setup for each round of our experiments is the following: input image size of $320$, $120000$ iterations with batch size $128$,
momentum SGD~\cite{pmlr-v28-sutskever13} optimizer with learning rate $1\mathrm{e}^{-3}$ divided by $10$ at iterations $80000$ and $100000$.
For the \textit{oracle}, we used the pretrained model EfficientDet D7~\footnote{
  \url{https://github.com/rwightman/efficientdet-pytorch} (Apache Licence 2.0).
}, 
with a score threshold of $0.6$.
For saving up computation, the round $0$ (i.e., first training before selection) was skipped in all experiments and replaced
by the loading of a model checkpoint trained with the same hyperparameters.
The annotator has close to state-of-the-art precision but is not perfect, and we observed in experiments cases~\footnote{
  this occurred on around $0.03$\% of annotated images
}
of false positive (e.g., a cropped truck that was wrongly identified as a bus), that strongly impacted the trained model accuracy.
Dealing with noisy labeled data is beyond the scope of this work, so we manually removed the experiments containing this kind of annotation.

As images with no bounding box annotation are ignored in our detector's loss function, we filtered out images from $S_n$ when
the oracle detects no object. Also, if after filtering, no image remains (i.e., $|S_n| = 0$), then $D_n = D_{n -1}$, so we skipped the training of round $n$.

In this setting, the elapsed time during selection was negligible compared to the training time, and training a model took around $7$ hours,
so a single experiment of active learning took up to $70$ hours.

\section{More Experimental Results}
\label{appendix:more_exp}

\subsection{More classification results}
\begin{figure}[!t]
  \centering
  \begin{minipage}[t]{0.49\linewidth}
    \centering
    \includegraphics[keepaspectratio, width=\textwidth]{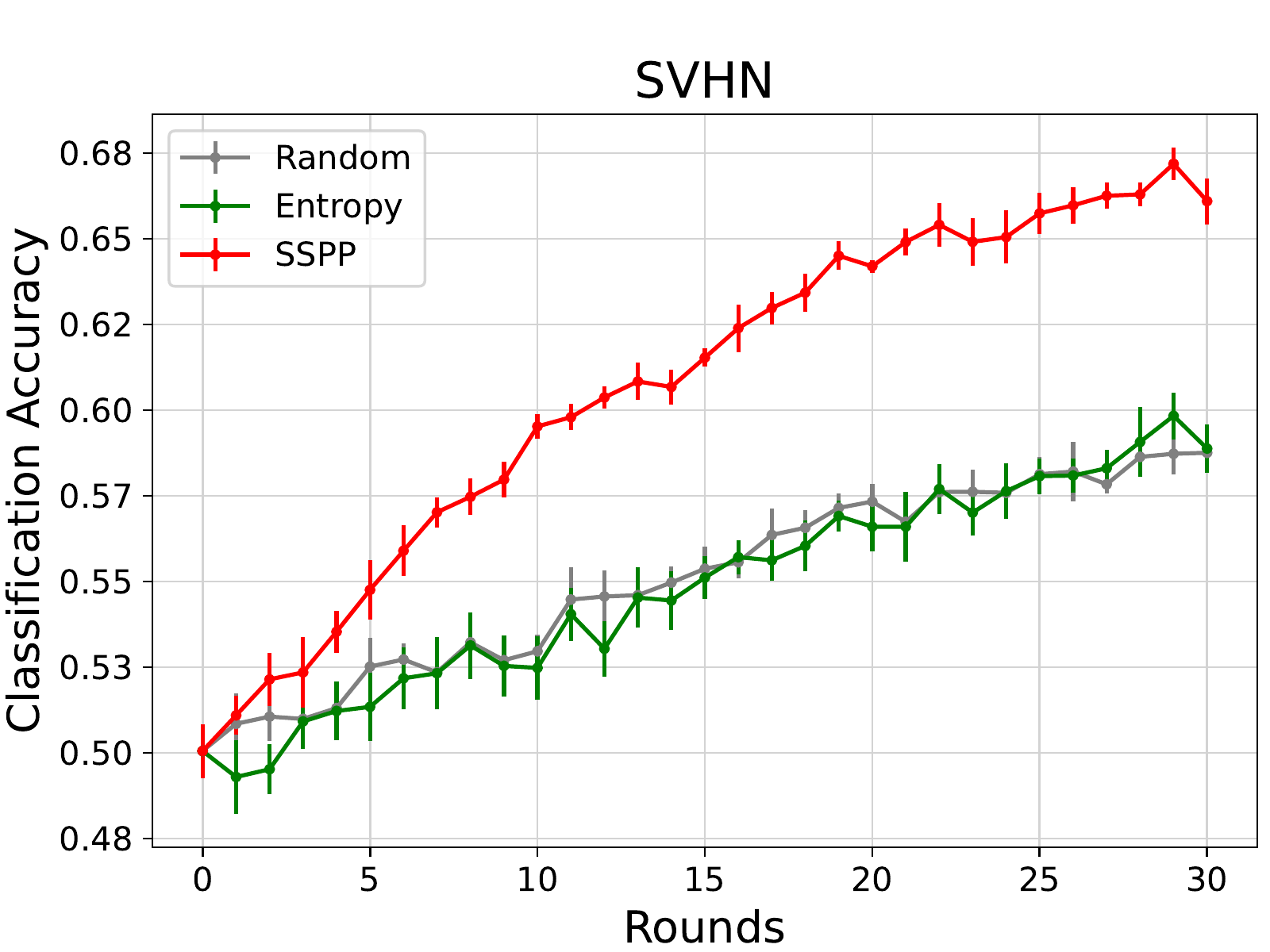}
    \subcaption{Cold-start scenario}
    \label{fig:classification_results_cs_appendix}
  \end{minipage}
  \begin{minipage}[t]{0.49\linewidth}
    \centering
    \includegraphics[keepaspectratio, width=\textwidth]{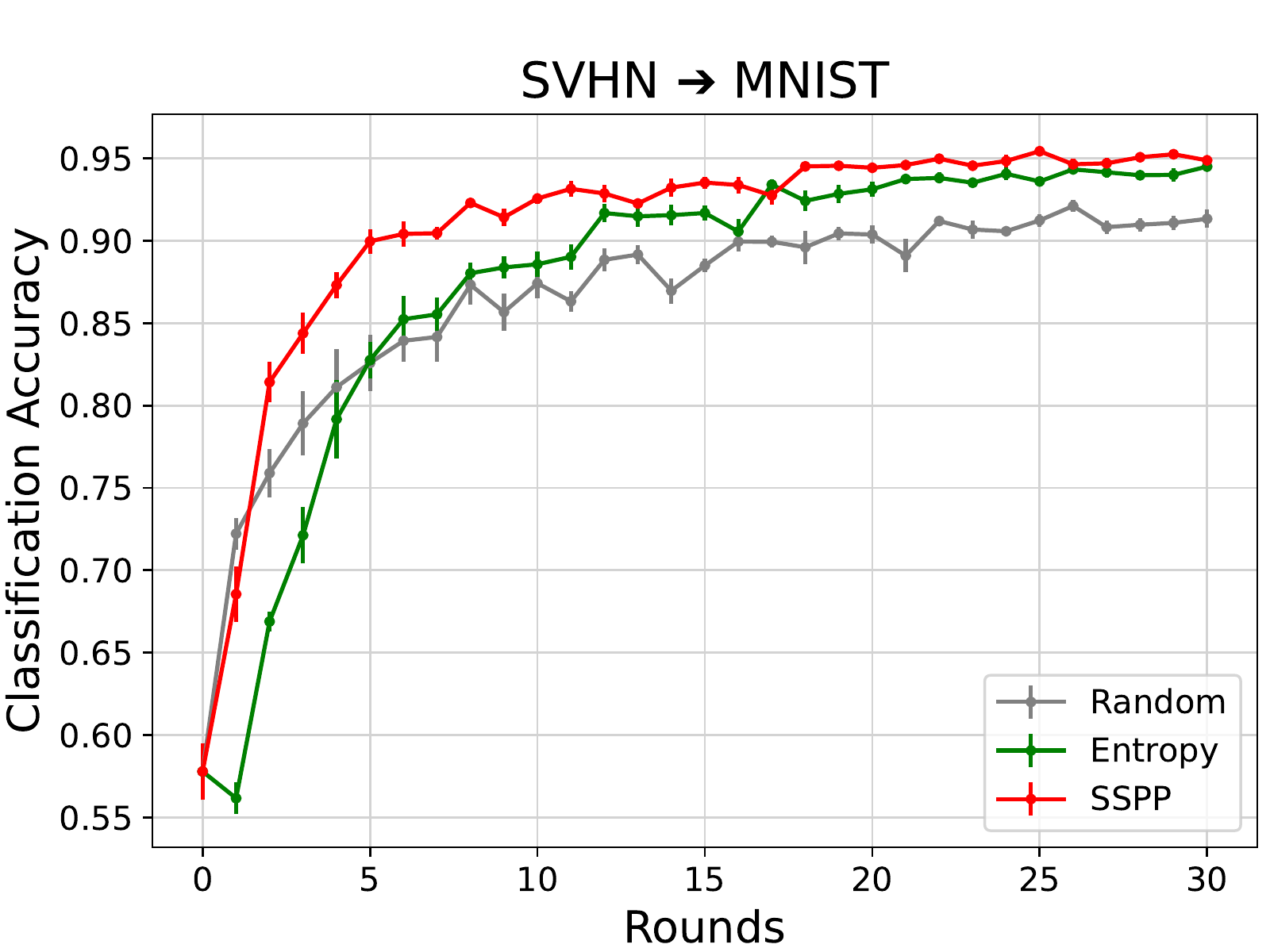}
    \subcaption{Domain adaptation scenario}
    \label{fig:classification_results_da_appendix}
  \end{minipage}
  \caption{
    Results of classification experiment
    in \subref{fig:classification_results_cs} the cold-start scenario
    and \subref{fig:classification_results_da} the domain adaptation scenario
    (mean accuracy vs. round number).
    Vertical bars show the std. err. across 5 repeats of the experiment
    with different random seeds.
  }
\end{figure}

We show the result of classification experiments in cold-start scenario with SVHN (Figure~\ref{fig:classification_results_cs_appendix}) and
in domain adaptation with SVHN $\rightarrow$ MNIST (Figure~\ref{fig:classification_results_da_appendix}).
In both experiments, the results showed the same tendency as in the other experiments described in Section~\ref{sec:experiments}.

\subsection{Number of Selected Images}
\begin{table*}[!t]
  \caption{The total number of unique and selected images in the classification experiment.
           Each value is the mean and std. err. across 5 experiments repeated
           with different random seeds.}
  \label{table:selected_samples}
  \small
  \begin{minipage}[t]{\linewidth}
    \subcaption{Cold-start scenario}
    \centering
      \begin{tabular}{ccccccc}
        \hline
        {} & \multicolumn{2}{c}{MNIST} & \multicolumn{2}{c}{SVHN} & \multicolumn{2}{c}{CIFAR-10} \\
        {} & \#unique & \#selected & \#unique & \#selected & \#unique & \#selected \\ \hline
        \texttt{Random} & 1074{\tiny $\pm$8} & 3840{\tiny $\pm$0} & 1074{\tiny $\pm$8} & 3840{\tiny $\pm$0} & 1074{\tiny $\pm$8} & 3840{\tiny $\pm$0} \\
        \texttt{Entropy} & 753{\tiny $\pm$2} & 3840{\tiny $\pm$0} & 831{\tiny $\pm$6} & 3840{\tiny $\pm$0} & 838{\tiny $\pm$3} & 3840{\tiny $\pm$0} \\
        \texttt{SSPP} & 2638{\tiny $\pm$23} & 3796{\tiny $\pm$3} & 2783{\tiny $\pm$30} & 3812{\tiny $\pm$2} & 2657{\tiny $\pm$25} & 3816{\tiny $\pm$3} \\ \hline
      \end{tabular}
    \label{table:classification_cs}
  \end{minipage}
  \vskip 0.15in
  \begin{minipage}[t]{\linewidth}
    \subcaption{Domain adaptation scenario}
    \centering
        \begin{tabular}{ccccccc}
        \hline
        {} & \multicolumn{2}{c}{SVHN $\rightarrow$ MNIST} & \multicolumn{2}{c}{MNIST $\rightarrow$ SVHN} & \multicolumn{2}{c}{STL-10 $\rightarrow$ CIFAR-10} \\
        {} & \#unique & \#selected & \#unique & \#selected & \#unique & \#selected \\ \hline
        \texttt{Random} & 1074{\tiny $\pm$8} & 3840{\tiny $\pm$0} & 1074{\tiny $\pm$8} & 3840{\tiny $\pm$0} & 1074{\tiny $\pm$8} & 3840{\tiny $\pm$0} \\
        \texttt{Entropy} & 551{\tiny $\pm$7} & 3840{\tiny $\pm$0} & 843{\tiny $\pm$10} & 3840{\tiny $\pm$0} & 882{\tiny $\pm$6} & 3840{\tiny $\pm$0} \\
        \texttt{SSPP} & 3315{\tiny $\pm$8} & 3829{\tiny $\pm$1} & 2851{\tiny $\pm$9} & 3822{\tiny $\pm$2} & 3353{\tiny $\pm$16} & 3829{\tiny $\pm$1} \\ \hline
      \end{tabular}
    \label{table:classification_da}
  \end{minipage}
\end{table*}

We report the number of selected images in classification experiments (Section~\ref{sec:experiments}) in Table~\ref{table:selected_samples}.
In this experiment, the images in a stream are duplicated,
and the number of unique images is a good indicator to see how diverse were the batches selected.
Moreover, we show the total number of selected images,
since \texttt{SSPP} may select fewer images than allowed
($128 \times 30 = 3840$ in this experiments).
While \texttt{Random} and \texttt{Entropy} always select the maximum number of images,
we can see that \texttt{SSPP} selects fewer.
This could have some impact on performance.

 \subsection{Study of $\epsilon$ Parameter in \SSPP}

\begin{figure}[!t]
\centering
\includegraphics[keepaspectratio, width=.3\textwidth]{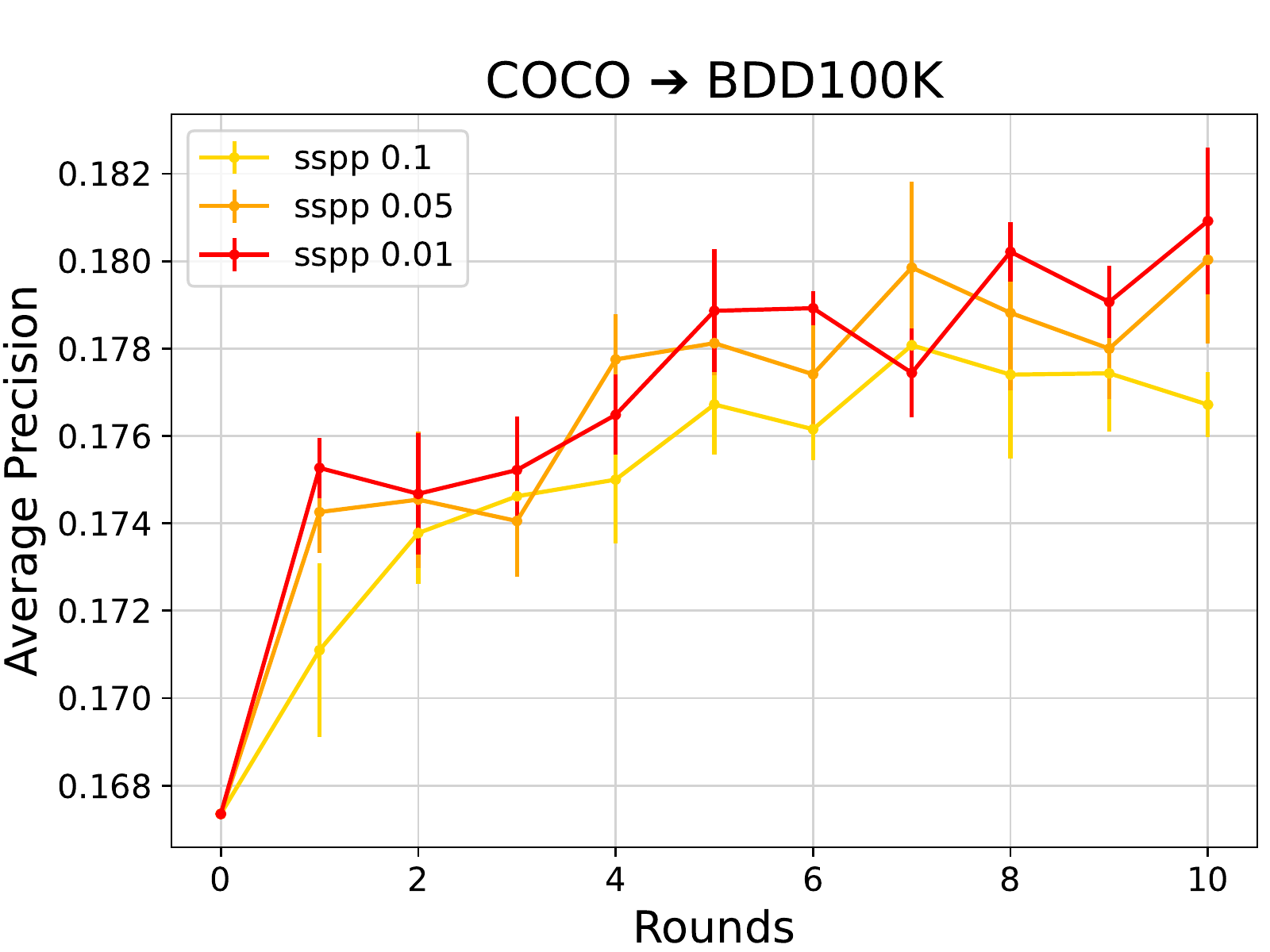}
\caption{
Results of domain adaptation experiment on \SSPP for different values of $\epsilon$ (average precision vs. round number). Vertical bars show the standard error.
}
\label{fig:detection_results_da1}
\end{figure}

As discussed in Section~\ref{sec:related_work}, the $\epsilon$ parameter of \SSPP allows a tradeoff between the complexity of the algorithm and the optimality of the solution.
We thus report in Figure~\ref{fig:detection_results_da1} the performance of the method depending on the value of $\epsilon$.
Conform to our expectations, the higher the $\epsilon$, the lower the performance.
This is consistent with the approximation ratio of the algorithm ($1/2 - \epsilon$).
However, the performance gain between $0.05$ and $0.01$ is small.
This shows signs of plateauing for smaller values of epsilon in this setting,
while the difference of computation cost is still large.
Based on this result, and the fact the user has a minimum requirement in terms of execution speed, the procedure for determining the value of $\epsilon$ in practice could be as follows:
since there is a value of $\epsilon$ at which the improvement in accuracy reaches a ceiling,
if one can afford to search this value through preliminary experiments (i.e., by performing labeling and training), 
the final $\epsilon$ is chosen by considering it, within the range of values that allows practical execution speed.
Otherwise, measuring the execution speed with several $\epsilon$ can be done at almost no cost, so one may just use the smallest $\epsilon$ fulfilling the speed requirements.

\subsection{Examples of Selected Samples}
\begin{figure*}[!t]
  \centering
  \includegraphics[keepaspectratio, width=1\textwidth]{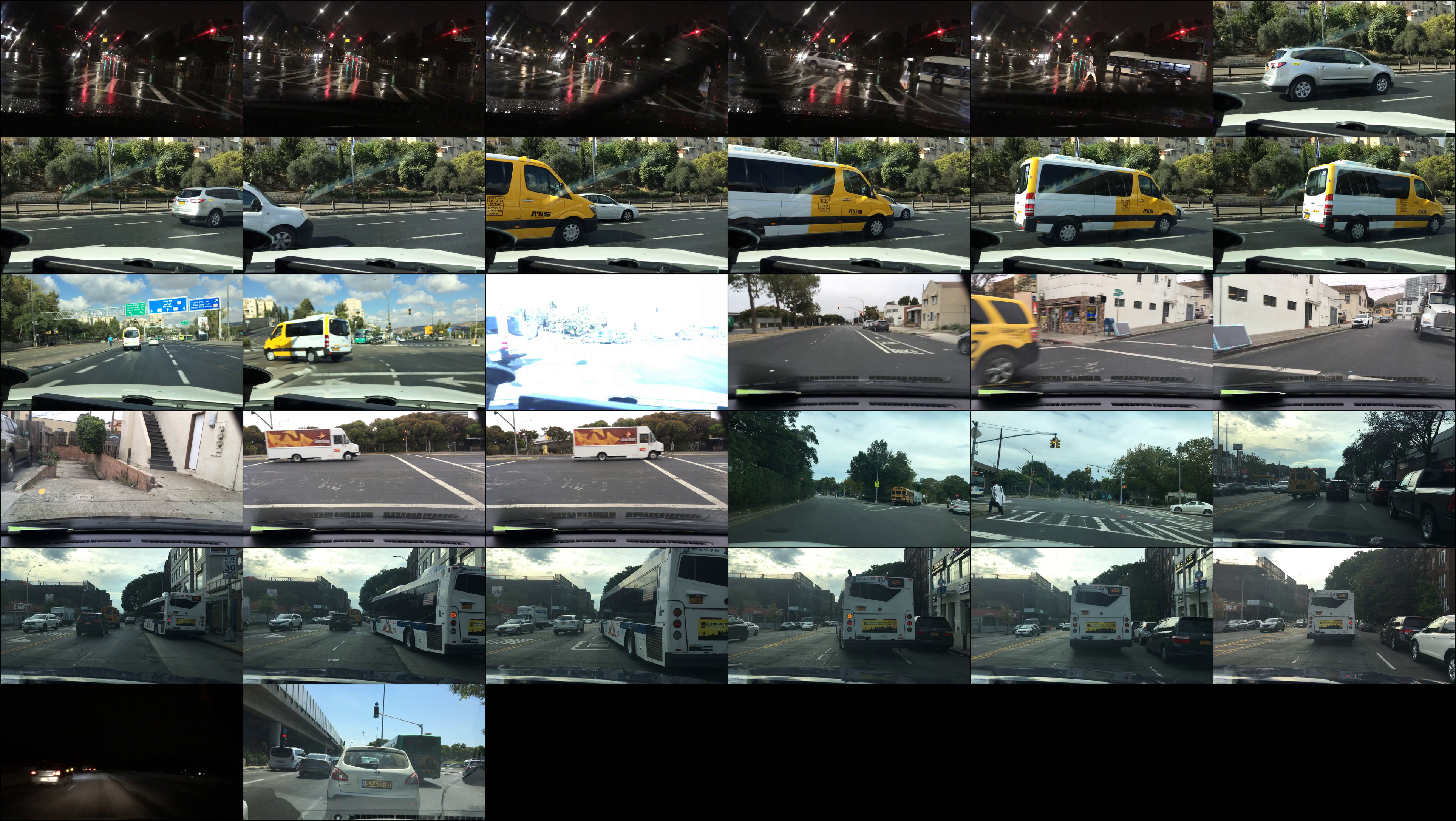}
  \caption{
    Batch of samples selected by \SSPP algorithm.
  }
  \label{fig:sample_sspp}
\end{figure*}

\begin{figure*}[!t]
  \centering
  \includegraphics[keepaspectratio, width=1\textwidth]{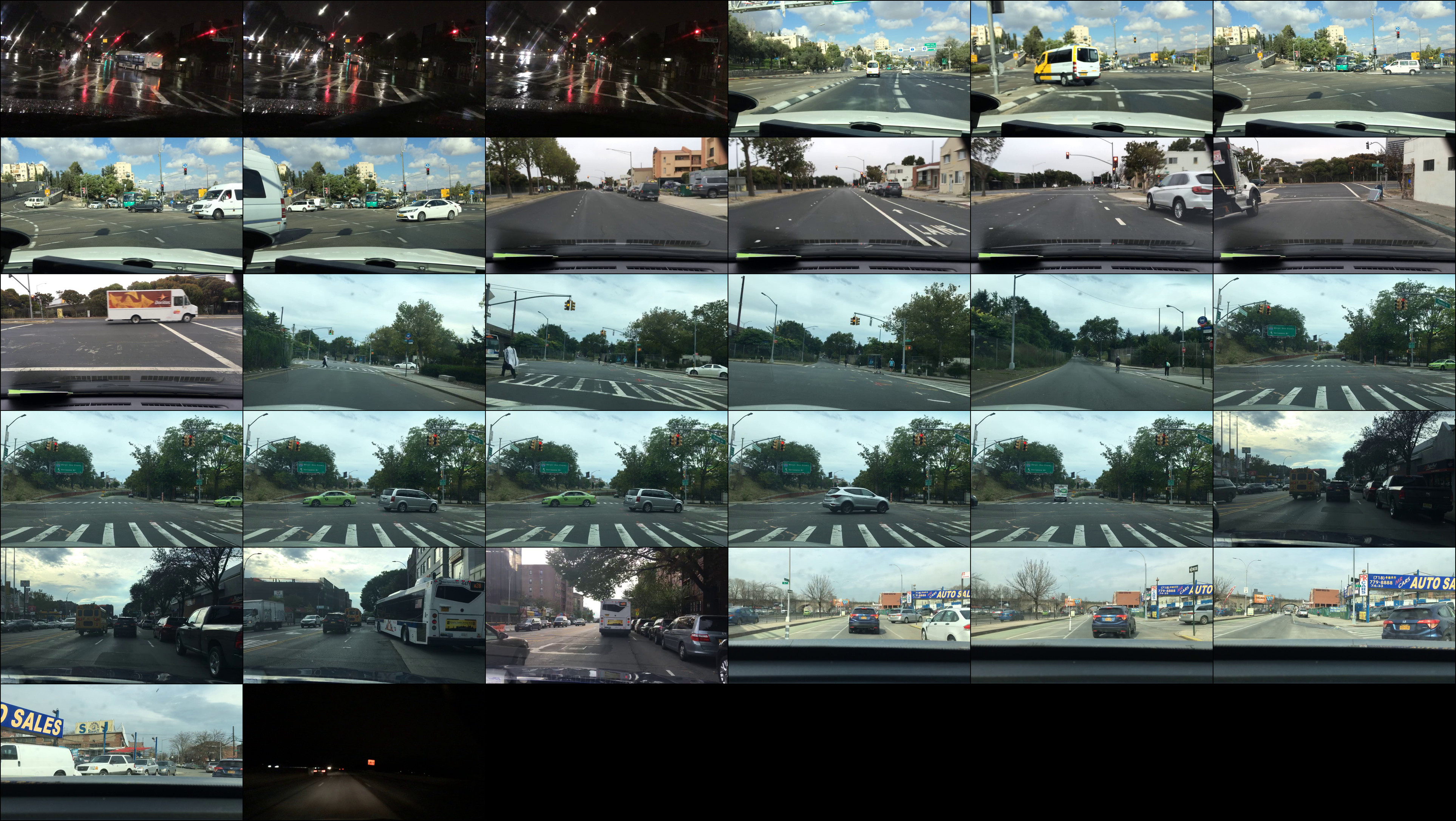}
  \caption{
    Batch of samples selected by random algorithm.
  }
  \label{fig:sample_random}
\end{figure*}

\begin{figure*}[!t]
  \centering
  \includegraphics[keepaspectratio, width=1\textwidth]{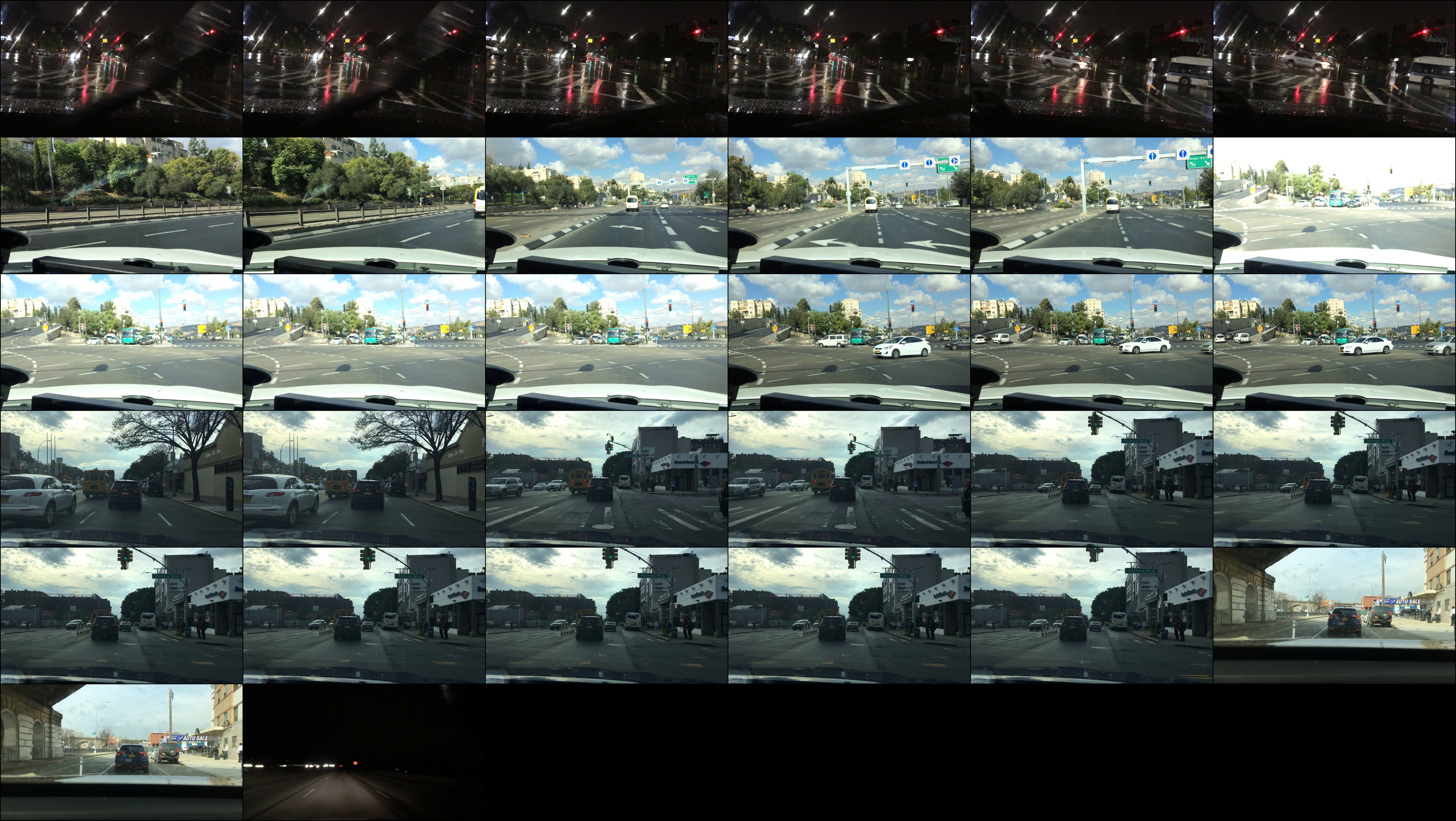}
  \caption{
    Batch of samples selected with only informativeness criteria (no diversity).
  }
  \label{fig:sample_maxunc}
\end{figure*}

We show the batches of images selected by different methods on the same round in
Figure~\ref{fig:sample_sspp},~\ref{fig:sample_random},~\ref{fig:sample_maxunc}.
The batch selected by \texttt{Entropy} in Figure~\ref{fig:sample_maxunc} contains a lot of informative samples,
as we can see instances hard to detect (e.g., bus located far from the camera) or close to decision boundaries (e.g., trucks looking like buses).
The redundance of samples in the batch is the consequence of the absence of a diversity component.
The batch selected by \texttt{SSPP} in Figure~\ref{fig:sample_sspp} shows a good tradeoff between informative samples with diversity.

\end{document}